\documentclass[letterpaper, 10 pt, conference]{ieeeconf}
\usepackage{graphicx}
\usepackage{amsmath,amssymb} % define this before the line numbering.
\let\labelindent\relax 
\usepackage{enumerate}
\usepackage[shortlabels]{enumitem} 
\usepackage{color}
\usepackage{tikz}
\usepackage{booktabs}
\usepackage{multirow}
\usepackage{authblk} 
\makeatletter
\let\NAT@parse\undefined
\makeatother
\usepackage{tikz} 
\usepackage{hyperref}
\hypersetup{
    colorlinks=true,
    linkcolor=blue,
    filecolor=magenta,  
    citecolor = blue,    
    urlcolor=cyan,
}

\graphicspath{{images/}} 
\begin{document}
\pagestyle{headings}
%\mainmatter

\newcommand{\cO}{\mathcal{O}}
\newcommand{\cG}{\mathcal{G}}
\newcommand{\cL}{\mathcal{L}}
\newcommand{\cc}{\mathcal{C}}

\newcommand{\bx}{\mathbf{x}}
\newcommand{\cX}{\mathcal{X}}
\newcommand{\cI}{\mathcal{I}}
\newcommand{\cJ}{\mathcal{J}}
\newcommand{\cD}{\mathcal{D}}
\newcommand{\cE}{\mathcal{E}}
\newcommand{\cA}{\mathcal{A}}
\newcommand{\bP}{\mathbf{P}}
\newcommand{\bbR}{\mathbb{R}}
\newcommand{\bp}{\mathbf{p}}
\newcommand{\bu}{\mathbf{u}}
\newcommand{\bv}{\mathbf{v}}
\newcommand{\ba}{\mathbf{a}}
\newcommand{\bb}{\mathbf{b}}
\newcommand{\bc}{\mathbf{c}}
\newcommand{\bl}{\mathbf{l}}
\newcommand{\bue}{\uvec{e}}
\newcommand{\buu}{\uvec{u}}
\newcommand{\bs}{\mathbf{s}}
\newcommand{\bn}{\mathbf{n}}
\newcommand{\bun}{\uvec{n}}
\newcommand{\bunu}{\uvec{$\nu$}}
\newcommand{\br}{\mathbf{r}}
\newcommand{\bpn}{\mathbf{pn}}
\newcommand{\bzero}{\mathbf{0}}
     
\newcommand{\btheta}{\bm{\theta}}
\newcommand{\bvartheta}{\bm{\vartheta}}
\newcommand{\inv}{^{\raisebox{.2ex}{$\scriptscriptstyle-\!1$}}}

% \def\ACCV18SubNumber{617}  % Insert your submission number here

% \accvfinalcopy 
%===========================================================
%\title{Disentangle Semantic labels: the case of occluded regions} % Replace with your title

\IEEEoverridecommandlockouts                              % This command is only needed if 
                                                          % you want to use the \thanks command

\overrideIEEEmargins                                      % Needed to meet printer requirements.

\title{Seeing Behind Things: \\ Extending Semantic Segmentation to Occluded Regions} % Replace with your title
%\titlerunning{Seeing behind things}
%\authorrunning{Purkait and Zach}

\author[1]{Pulak Purkait$^*$\thanks{$*$The work was partially done at Toshiba Research Europe, Cambridge}}
\author[2]{Christopher Zach}
\author[1]{Ian Reid}
\affil[1]{The University of Adelaide, Australia}
\affil[2]{Chalmers University of Technology, Sweden}
%\institute{Toshiba Research Europe Ltd., Cambridge}

\maketitle

%===========================================================
\begin{abstract}
  Semantic segmentation and instance level segmentation made substantial
  progress in recent years due to the emergence of deep neural networks
  (DNNs). A number of deep architectures with Convolution Neural Networks
  (CNNs) were proposed that surpass the traditional machine learning
  approaches for segmentation by a large margin. These architectures predict
  the directly \emph{observable} semantic category of each pixel by
  usually optimizing a cross-entropy loss. In this work we push the limit of
  semantic segmentation towards predicting semantic labels of directly visible
  as well as occluded objects or objects parts, where the network's input is a
  single depth image. We group the semantic categories into one
  background and multiple foreground object groups, and we propose a
  modification of the standard cross-entropy loss to cope with the settings. In our
  experiments we demonstrate that a CNN trained by minimizing the proposed loss is able to predict semantic
  categories for visible and occluded object parts without requiring to
  increase the network size (compared to a standard segmentation task). The
  results are validated on a newly generated dataset (augmented 
  from SUNCG) dataset.
\end{abstract}

%-------------------------------------------------------------------------
\section{Introduction}
\label{sec:intro}

Semantic image segmentation is one of the standard low-level vision tasks in
image and scene understanding: a given RGB (or depth or RGB-D) image is
partitioned into regions belonging to one of a predefined set of semantic
categories. The focus of standard semantic segmentation is to assign semantic
labels only to directly observed object parts, and consequently reasoning
about hidden objects parts is neglected. Humans (and likely other sufficiently
advanced animals) are able to intuitively and immediately ``hallucinate''
beyond the directly visible object parts, e.g.\ human have no difficulties in
predicting wall or floor surfaces even when they are occluded in the current
view by cupboards or desks.

In this work we propose to extend the traditional semantic segmentation
problem---which assigns exactly one semantic label per pixel---to a
segmentation task returning a set of semantic labels being present (directly
visible or hidden) at each pixel. Thus, the predicted output of our approach
is the semantic category of visible surfaces as well as the likely semantic
labels of occluded surfaces. Since the underlying task is an ill-posed
problem, at this point we allow two simplifications: first, we make this
problem easier by grouping finer-grained semantic categories into coarser
semantic groups; and second, we leverage strong supervision for learning by
relying on synthetic training data generation.

\begin{figure}
\centering \scriptsize 
{\includegraphics[width=0.4\textwidth]{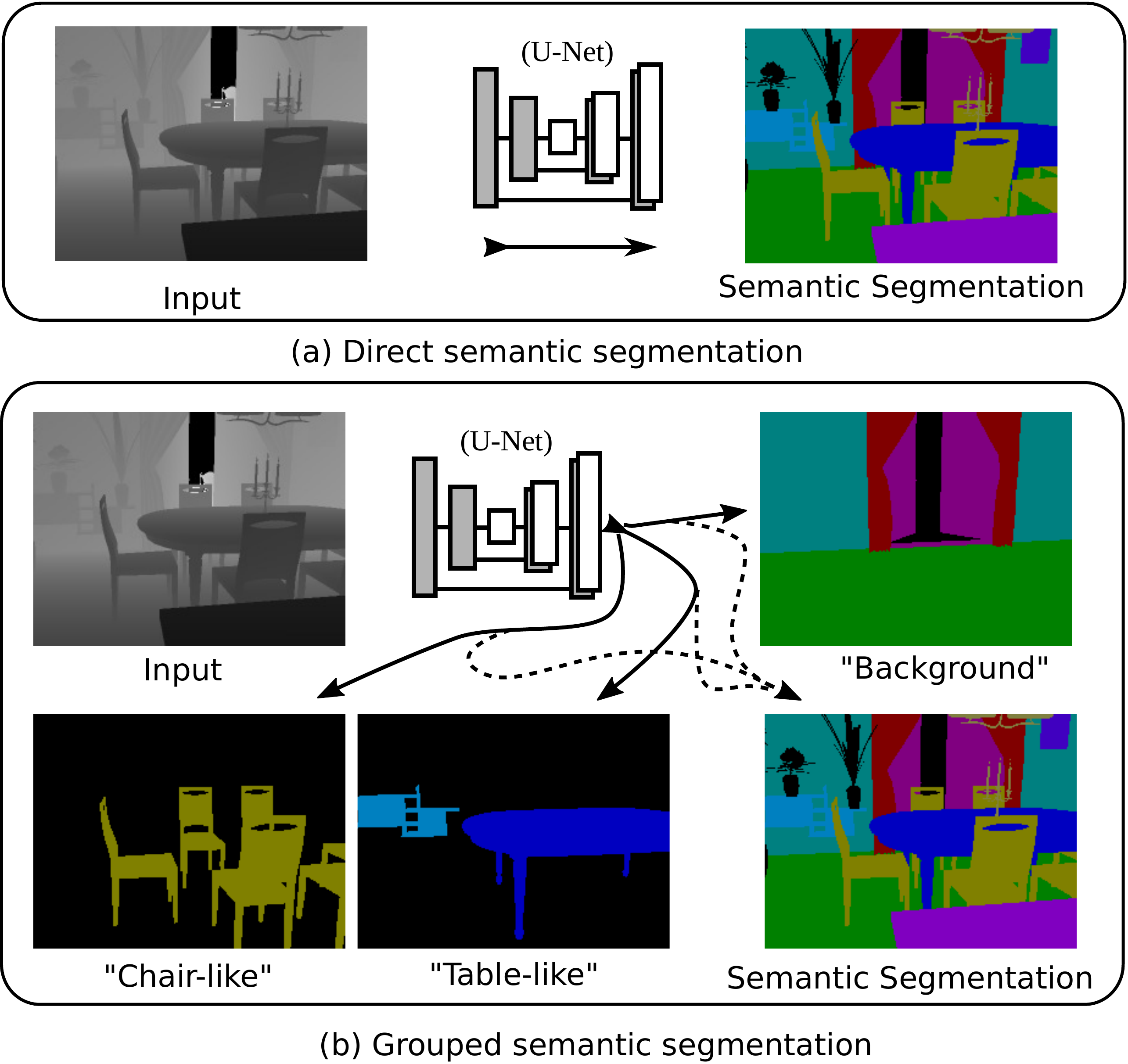}} 
\caption{Conventional semantic segmentation vs Our group-wise semantic segmentation. The proposed grouped semantic segmentation method estimates labels of the visible objects from each group and also the visibility of the individual groups which could be further  merged together to get the semantic labels. Thus, our method generates more information that an agent could utilize for indoor / outdoor navigation. (See the results in Fig.~\ref{fig:SUNCGResults})} 
\label{fig:teaser}
\end{figure}

To our knowledge there is relatively little literature aiming to predict
properties of hidden object parts such as extending the semantic category to
unseen surfaces. There is a large body of literature on standard semantic
image segmentation, and we refer to~\cite{garcia2017review} for a relatively
recent survey. If we go back to classical object detection methods estimating
bounding boxes (such as~\cite{redmon2016_YOLO}), one can argue that such
multi-class object detection approaches already provide a coarse idea on the
total extent of a partially occluded object. This observation was e.g.\ used
in~\cite{yang2010layered} to reason about layered and depth-ordered image
segmentation. A more refined approach to bounding-box based object detection
is semantic instance segmentation
(e.g.~\cite{hariharan2014simultaneous,dai2016instance,he2017mask,kirillov2018panoptic}), which
provide finer-grained cues to reason about occluded object parts. We bypass
explicit object detection or instance segmentation by directly predicting
labels for all categories present (visible or occluded) at each pixel.

This work is closely related to {\bf amodal segmentation}
works~\cite{li2016amodal,zhu2017semantic,ehsani2018segan}. Generally, amodal perception refers
to the intrinsic ability of humans to perceive objects as complete even if
they are only partially visible. We refer to~\cite{breckon2005amodal} for an
introduction and discussion related to (3D) computer vision.  Our work differs
from earlier works~\cite{li2016amodal,zhu2017semantic} as our approach
predicts amodal segmentation masks for the entire input image (instead of
predicting masks only for individual foreground objects). We also target
amodal segmentation for rich environments such as indoor scenes and urban
driving datasets. We also leverage synthetic training data generation instead
of imprecise human annotations.

More recently, there are several attempts to use deep learning to either
complete shapes in 3D from single depth or RGB images
(e.g.~\cite{wu2015_3dshapenets,sharma2016vconv}) or to estimate a full
semantic voxel space from an RGB-D image~\cite{song2017semantic}. The
utilization of a volumetric data structure enables a very rich output
representation, but severely limits the amount of detail that can be
estimated. The output voxel space used in~\cite{song2017semantic} has a
resolution of about $50^3$ voxels. The use of 3D voxel space (for input and
output) and 3D convolutions also limits the processing efficiency. The
low-resolution output space can be addressed e.g.\ by multi-resolution
frameworks (such as~\cite{dai2018scancomplete}), but we believe that
image-space methods are more suitable (and possibly more biologically
plausible) than object-space approaches.

In summary, our contributions in this work are as follows:
\begin{itemize}[noitemsep,nolistsep,leftmargin=5.5mm,topsep=1ex,partopsep=1ex] 
\item We propose a new method for semantic segmentation going beyond assigning
  categories for visible surfaces, but also estimating semantic segmentation
  for occluded objects and object parts.
\item Using the SUNCG dataset we create a new dataset to train deep networks
  for the above-mentioned segmentation task.
\item We show experimentally that a CNN architecture for regular semantic
  segmentation is also able to predict occluded surfaces without the necessity
  to enrich the network's capacity.
%\item A new strategy is proposed for semantic segmentation that can estimate semantic labels of the occluded object parts given a single RGB / depth image. 
%\item Under the proposed grouping strategy and group-wise semantic loss, our method follows similar quantitative matrices such as accuracy, execution time and memory footprint with the conventional cross entropy loss for semantic segmentation. However, unlike existing methods, our method can estimate labels of the occluded parts.  
\end{itemize}
The last item suggests that predicting the semantic labels of occluded parts
is as a classification problem not significantly more difficult than regular
semantic segmentation, and that there are likely synergies between estimating
object categories for directly visible and occluded parts.
%This seems plausible, since a network successfuly solving a standard semantic segmentation task will have a good internal represention of the spatial context at each pixel.
Hence, one of the main practical issues is the generation of real-world
training data, which we evade by leveraging synthetic scenes. A schematic 
comparison of our method against the traditional semantic segmentation 
is illustrated in Fig.~\ref{fig:teaser}.

%One might argue that the proposed method is similar to {\bf amodal instance segmentation}~\cite{li2016amodal}, \cite{zhu2017semantic}.
%The amodal perception is the perception of the whole of a physical structure when only parts of it are visible~\cite{breckon2005amodal}, thus our method could also be inferred as amodal segmentation.
%However, our methods defer from the above amodal segmentation methods~\cite{li2016amodal}, \cite{zhu2017semantic} in the following way: (i) instead of modeling individual objects and natural scenes---our target domain is an indoor environment / driving environment
%where the scene is constrained by a group of objects and their occurrences,(ii) the datasets are generated based on human perception which has high
%uncertainty of the ground-truth labels across different subjects---in contrast, our datasets (based on SUNCG~\cite{song2017semantic}) are generated
%by rendering full 3D models of different objects and hence cleaner ground-truth.

This paper is organized as follows: section~\ref{sec:learning} formalizes the
overall problem and present our solution, and section~\ref{sec:evaluation}
describes the generation of the dataset and the obtained qualitative and
quantitative evaluation. Finally, section~\ref{sec:conclusion} concludes this
work and outlines future work.

%------------------------------------------------------------------------- 

\section{Learning to segment visible and hidden surfaces}
\label{sec:learning}

In this section we describe our proposed formulation in more detail. We start
with illustrating the semantic grouping strategy, which is followed by
specifying the utilized loss function and CNN architecture.

\subsection{Semantic grouping }\label{sec:grouping}
The proposed grouping strategy is intuitively based on the criterion that
objects in a particular group do not necessarily occlude another objects from
the same group. On the other hand objects from separate groups can occlude
each other. Examples of such grouping can be found in section
\ref{suncgdatasets}.  For example the ``Chair-like'' group in Table~\ref{tab:groupSUNCG} contains similar looking objects 'chair',
'table$\_$and$\_$chair', 'trash$\_$can', and 'toilet' from living room, dining
room, kitchen and toilet respectively. Thus each of those objects usually do
not occur simultaneously in a particular image.  Note that multiple
occurrences of a particular object may occlude each other and such self-occlusions are not incorporated. At this point we rather focus on inter-group object category occlusions.  
 
Let there be $N$ different objects categories
$[\cO_1,\;\cO_2,\;\ldots,\cO_N]$ in the scene.  The task of semantic
segmentation is to classify the pixels of an image to one of those $N$ object
categories.  A pixel $p$ is marked as a particular category $\cO_j$ if the
posterior probability of the object category $\cO_j$ is maximum at the pixel
$p$. A straight-forward cross-entropy loss is generally utilized in the
literature for the said task. As discussed before, the current loss is limited
to classifying directly visible objects, i.e., if an object category is
visible in a particular pixel, we enforce posterior probability to be
$1$ at that pixel else $0$ through cross-entropy loss. In this work, we push
it further and allow multiple semantic labels to be active for each pixel. The
output of the trained classifier also indicates which of these active labels is
corresponding to the directly visible category, which therefore yields the
standard semantic segmentation output.  However, we do not directly infer a
full depth ordering of objects and their semantic categories (which we leave as future work). Rather, we
classify each pixel into one of $M$ different groups, where the grouping is
based on usual visibility relations. Each group contains a number of
finer-grained object categories in addition to a ``void'' category indicating
absence of this group category. Therefore, a pixel will be ideally assigned to
a visible and also all occluded object categories.
  
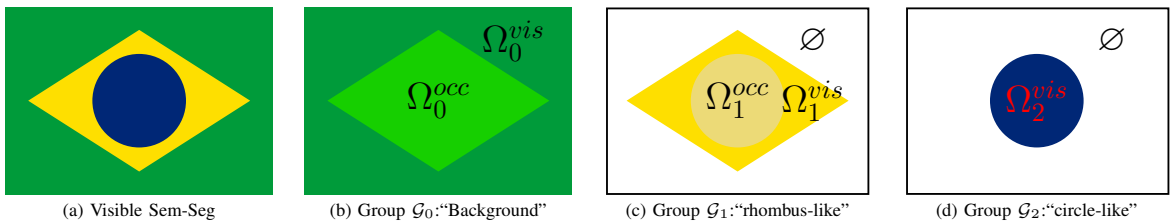
\begin{figure*}
\centering \scriptsize 
\begin{tabular}{cccc}
\includegraphics[width=0.2\textwidth]{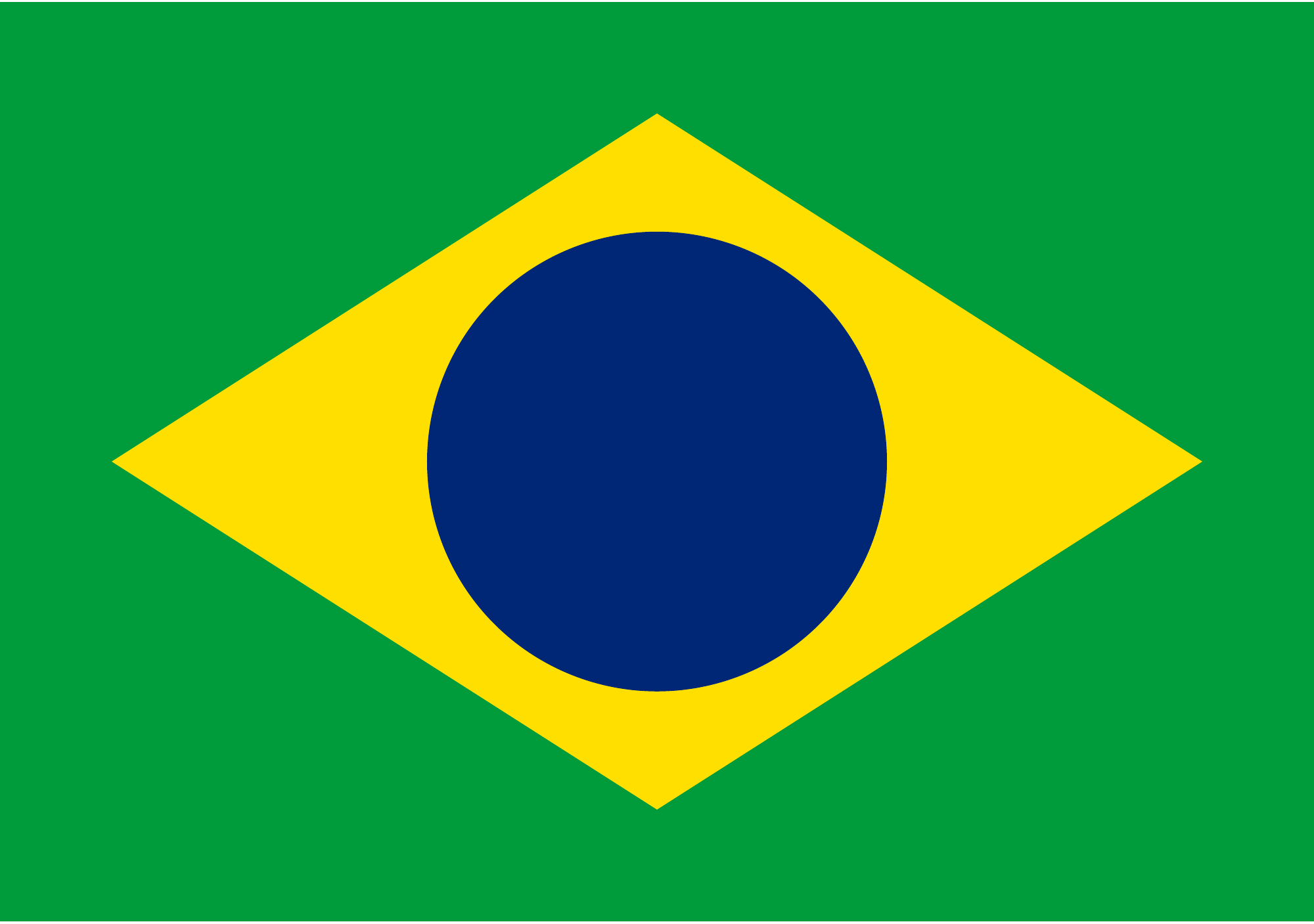}&
\begin{tikzpicture}
    \draw (0, 0) node[inner sep=0] {\includegraphics[width=0.2\textwidth]{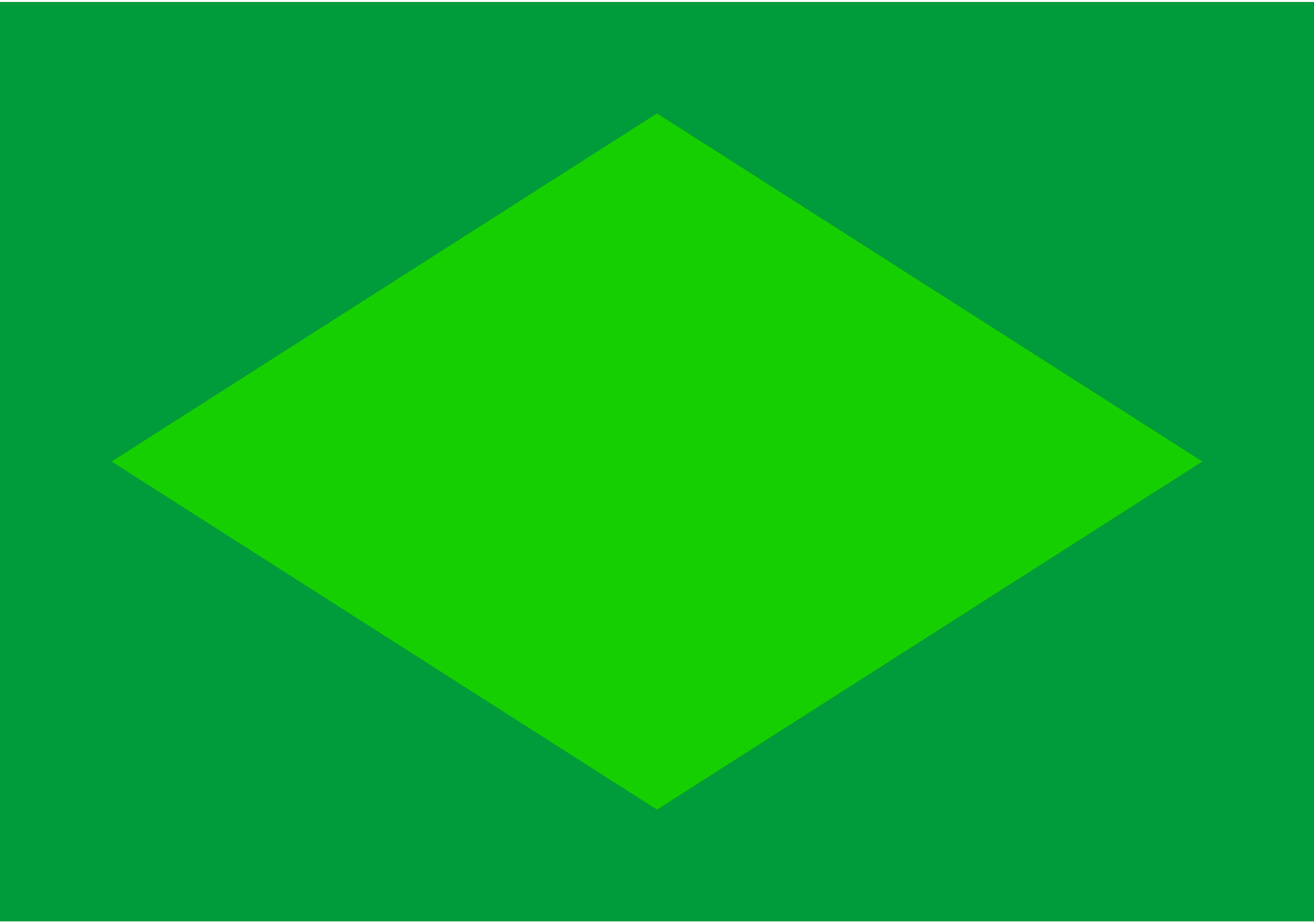}};
    \draw (1, 0.8) node {\Large $\Omega_0^{vis}$};
    \draw (0, 0.0) node {\Large $\Omega_0^{occ}$};
\end{tikzpicture} &
\begin{tikzpicture}
    \draw (0, 0) node[inner sep=0] {\includegraphics[width=0.2\textwidth]{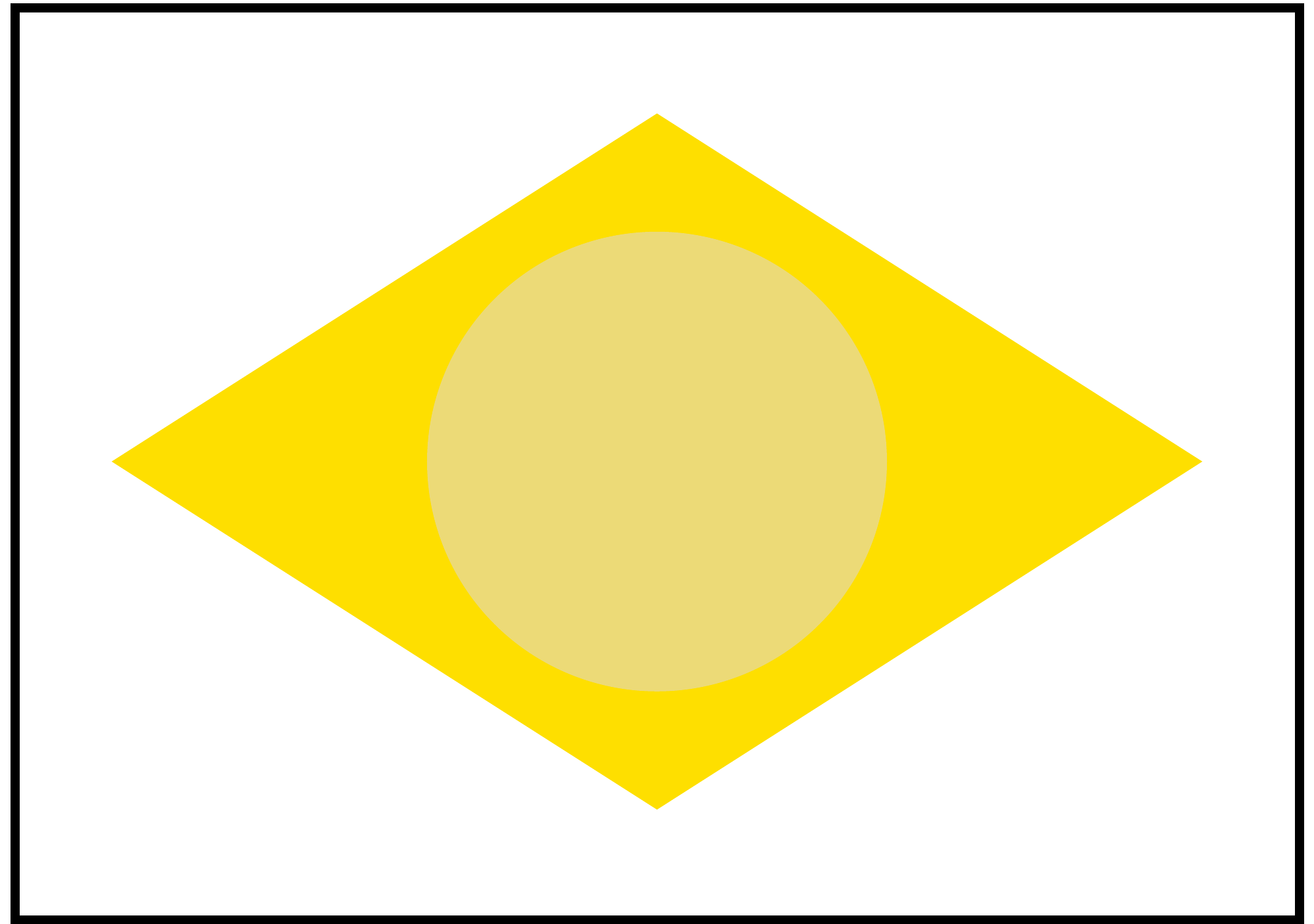}};
    \draw (1, 0.8) node {\Large $\varnothing$};
    \draw (0, 0.0) node {\Large $\Omega_1^{occ}$};
    \draw (1, 0.0) node {\Large $\Omega_1^{vis}$};
\end{tikzpicture} &
\begin{tikzpicture}
    \draw (0, 0) node[inner sep=0] {\includegraphics[width=0.2\textwidth]{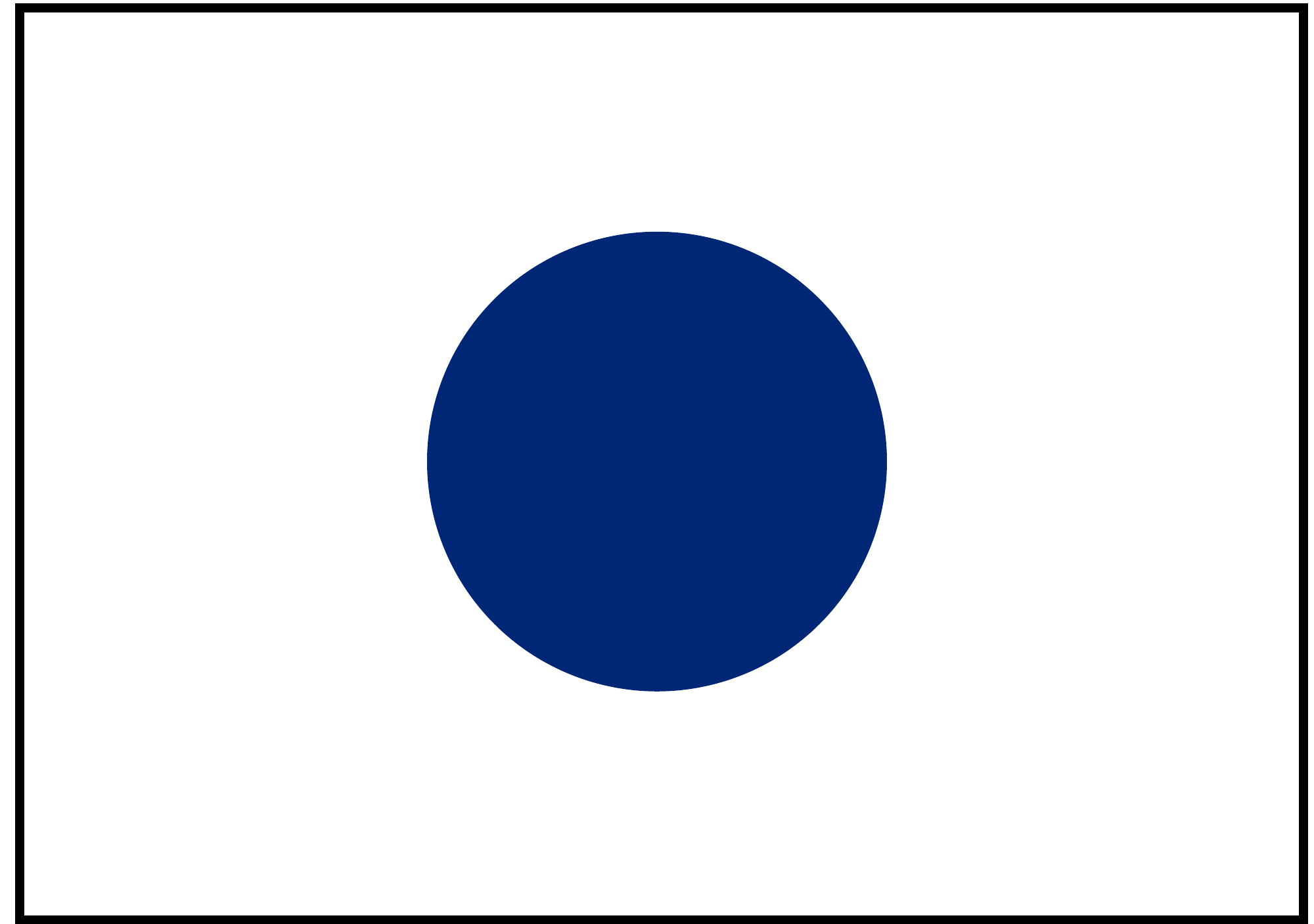}};
    \draw (1, 0.8) node {\Large $\varnothing$};
    \draw (0, 0.0) node {\Large {\color{red}$\Omega_2^{vis}$}};
\end{tikzpicture} \\ 
(a) Visible Sem-Seg & (b) Group $\cG_0$:``Background" & (c) Group $\cG_1$:``rhombus-like" & (d) Group $\cG_2$:``circle-like" 
\end{tabular}
\caption{An example of the group-wise visibility of four different groups is displayed. A particular color code represent an object category present in a group.  
The ``void'' category $\varnothing$ is not marked with any color. $\Omega_c^{vis}$ is marked with 
a darker color whereas $\Omega_c^{occ}$ is marked with a lighter color. Note that $\Omega_c^{pres} = \Omega_c^{vis} \cup \Omega_c^{occ}$. }
\label{fig:grouping}
\end{figure*}

We group $N$ objects into $M+1$ different groups
$[\cG_0,\;\cG_1,\;\cG_2,\;\ldots,\cG_M]$ where the group $\cG_i$ contains
$g_i = |\cG_i|$ object categories, \emph{i.e.},
$\sum_{i=0}^M g_i = N$. Note that $g_i$'s are not necessarily equal, and
different groups can have different number of object categories.  Our
assumption is that an objects category $\cO_{i_k} \in \cG_k$ do not occlude
another object category $\cO_{j_k} \in \cG_k$. The group $\cG_0$ is considered
as the back-ground.  We
extend $\cG_i$ by a ``void'' category $\varnothing$, yielding
$\cG_i := \cG_i \cup \{ \varnothing \}$ for $i=1,\dotsc,M$. The ``void'' category (with corresponding index 0) indicates absence of the group category at that pixel and is not used in the background group $\cG_0$. An example of such grouping is displayed in Fig.~\ref{fig:grouping}.

%\subsection{Pixel Classification Procedure / loss}
\subsection{Semantic segmentation output and loss}
\subsubsection{\bf Conventional Method}
The baseline methods for semantic segmentation compute pixel-wise soft-max
over the final feature map combined with the cross-entropy loss function. Note
that for an image of size $h \times w$, the network $f(.)$ produces feature
map of size $h \times w \times N$. The soft-max computes the posterior
probability of each pixel belonging to a particular class, which is computed
as
$p_c(x) = \exp(\alpha_c(x))/\left(\sum_{c^{\prime}=1}^N \exp(\alpha_{i^\prime}  (x))\right)$,
where $\alpha_c(x) = f_c(x;I)$ is the final layer activation in $i$th feature
channel (corresponding to $c$th class) at the pixel position $x$ for input
image $I$. In other words $(p_1(x), \dots p_N(x))$ is an element from the
$N$-dimensional probability simplex $\Delta^N$.  Note that number of
activations $N$ is the same as the number of object categories. The energy
function is defined by the multi-label cross-entropy loss function for
semantic segmentation as follows
\begin{align}
  \mathcal L_{CE} = \sum_{x \in \Omega} \log p_{c^*(x)}(x),
\end{align}
where $c^*(x)$ is the provided ground truth semantic label at pixel
$x$ and $\Omega$ is the image domain.
 
\subsubsection{\bf Proposed Method}

In our proposed formulation the classification output more structured than
above (where the output is an element from the $N$-dimensional probability
simplex). In our setting the prediction is an element from a product space
over probability simplices,
\begin{align}
  {\cal Y} := \Delta^{(M+1)} \times \prod\nolimits_{i=0}^M \Delta^{g_i+1},
\end{align}
where $(M+1)$ is the number of object groups, and $g_i = |\cG_i|$ is the number of
object categories in group $i$. We write an element from $\cal Y$ as
$(p, q^0, q^1, \dotsc, q^M)$, where $p \in \Delta^{(M+1)}$ and
$q^i \in \Delta^{g_i+1}$ for all $i=0,1,\dotsc,M$. We use an argument
$x$ to indicate the prediction at pixel position $x$. A perfect prediction
would have $p(x)$ and all $q_i(x)$ at the corner of the respective probability
simplex with $p_i(x)=1$ if the $i$-th subgroup is visible, and $q^i_j(x)=1$ for
$j\ge 1$ if the $j$-th object category in $\cG_i$ is present (directly visible
or occluded). If none of the categories in $\cG_i$ is present, then we read
$q^i_0(x) = 1$, i.e.\ index `$0$' of each group corresponds to the ``void'' category. 

One extreme case is if $|\cG_i|=1$, i.e.\ each group contains exactly one
object category. Then our representation amounts to independently predicting
object classes that are present ($q^i_1(x)=1$) and which of these object
categories is directly visible in the image at pixel $x$ (via
$p_i(x)$). We utilize larger groups for efficiency.

Further note that there are intrinsic constraints on element from
$\cal Y$ to be physically plausible. If a group $i$ is estimated to be visible
at pixel $x$ ($p_i(x) = 1$), then $q^i_0(x)$ has to be 0, as some category
from group $i$ is estimated to be observed directly. One can encode such
constraints as inequality constraints $p_i(x) \le 1-q^i_0(x)$ (or
$p_i(x) \le \sum_{j=1}^{g_i} q^i_j(x)$ as
$\sum_{j=0}^{g_i} q^i_j(x) = 1$). 
For simplicity we do not enforce this
constraints at test time and therefore rely on physically plausible training
data (always, $p_i(x) \le 1-q^i_0(x)$) in order for the network to predict only plausible outputs. 
Refer to sec.~\ref{sec:quantitative} for further analysis.

 % In this work, the loss function is defined on the group level. As mentioned above, instead of classifying 
 % a pixel to one of the $N$ object categories, we classify each pixel---(1) into one of $(M+1)$ different groups based on the visibility, and further, (2) for each group assign one of the object categories in the back-ground group $\cG_0$, and (3) also for each object group $\cG_k$ assign one of the object categories therein or back-ground. The number of  activations required with current representation---for group-wise visibility detection $(M+1)$,  for back-ground $g_0$ and for each group $(g_k+1)$. Therefore the total number of activations  $(M+1) + g_0 + \sum_{k=1}^M(g_k+1)$ compared to $N$ in the conventional representation. The difference 
 % $(M+1) + g_0 + \sum_{k=1}^M(g_k+1) - N = 2*M+1$ is rather small for a small number of groups. For example, if we model objects by just two different groups with foreground objects and background, the number of extra activation required is  $2*1 + 1=3$.
 
Note that standard semantic segmentation can also be inferred from our
group-wise segmentation, e.g.\ by assigning the semantic label corresponding
to $\max_i p_i(x) \max_{j=1,\dotsc,g_i} q^i_j(x)$. The details of the procedure is described in sec.~\ref{sec:quantitative}. 
In the experiment section, we show that we observe similar performance for the
task of semantic segmentation by a similar network trained with the
conventional method. Thus with a network architecture almost identical to one
for standard semantic segmentation, our proposed output representation
additionally estimates semantic labels of occluded back-ground and inter-group
occluded object categories with nearly the same number of parameters (only the output layer differs).

%\paragraph{\bf Modified loss function}
\subsubsection{\bf Modified loss function}

For a given image and ground truth semantic label $c^*(x)$ at pixel
$x$, let $i^*(x)$ be the semantic group such that
$c^*(x) \in \cG_{i^*(x)}$ and let $j^*(x) \ne 0$ be the (non-``void'') element in
$\cG_i$ corresponding to category $c^*(x)$. Let $\cc(i,j)$ be the reverse
mapping yielding the category corresponding to group $i$ and index $j$ (for
$j \ge 1$). Let $\Omega_c^{vis} \subseteq \Omega$ be the image regions labeled
with category $c$, i.e.\ pixels where category $c$ is directly
visible. Further we are given $N$ sets
$\Omega^{pres}_c \subseteq \Omega$ that indicate whether category $c$ is
present (directly visible or occluded) at a pixel. Now
$\Omega_c^{occ} = \Omega_c^{pres} \setminus \Omega_c^{vis}$ is
the set of pixels where category $c$ is present but not directly
visible. Finally,
$\Omega_i^\varnothing := \Omega \setminus \left( \bigcup_{j \in \cG_i}
  \Omega^{pres}_{c(i,j)} \right)$
is the set of pixels where no category in group $i$ is present.  With these
notations in place we can state the total loss function as
\begin{align}
  \cL &= \cL_g + \sum\nolimits_{i=0}^M \cL_i \label{eq:loss} 
\end{align}
with
\begin{align}
  \cL_g &= \sum_{x \in \Omega} \log p_{i^*(x)}(x) \\
  \cL_i &= \sum_{j \in \cG_i} \left( \sum_{x \in \Omega^{vis}_{c(i,j)}} \log q^i_j(x)
  + \lambda \sum_{x \in \Omega^{occ}_{c(i,j)}} \log q^i_j(x) \right) \\ \nonumber & ~~~~~~~~~~~~~~~~~~~~~~~~~~~ + \lambda \sum_{x \in \Omega_i^\varnothing} \log q^i_0(x).        
\end{align}
$\cL_g$ is a standard cross-entropy loss such that the correct visible group
is predicted. For each group $i$ we have a cross-entropy loss aiming to
predict the correct visible index $j$ in group $i$ (with full weight) and
occluded indices (with weighting $\lambda$).
Throughout all of our experiments the weight proportion $\lambda$ is chosen as $0.1$.

\subsection{Architecture}\label{sec:architecture}
We utilize a standard semantic segmentation architecture (U-Net~\cite{ronneberger2015u}) with a deep CNN. It 
consists of an encoder and a decoder which are sequences of convolutional layers followed by instance normalization~\cite{ulyanov2016instance} layers and ReLus except for  the last layer where a sigmoid activation function is utilized followed by group-wise soft-max. Skip connections from encoder to decoder are also incorporated in the architecture. %The details of network parameters and architecture can be found in the supplementary material. 

\begin{figure}
\centering \scriptsize 
\includegraphics[width=0.5\textwidth]{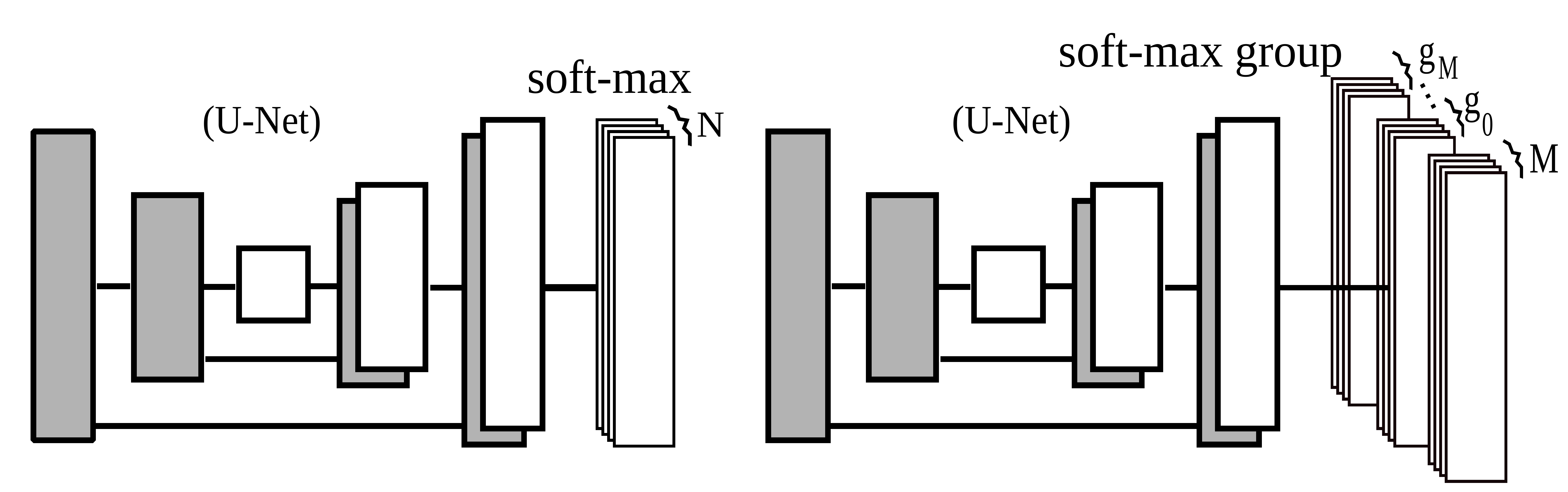} 
\begin{tabular}{cc}
 (a) Direct Sem-Seg ~~~~~~~~ & ~~~~(b) Grouped Sem-Seg
\end{tabular}
\caption{Conventional semantic segmentation vs Our group-wise semantic segmentation: (a) in the conventional semantic 
segmentation there are $N$ number of activations and (b) in the proposed method we have $2(M+1) + N$ activation layers. For details see the text.}
\label{fig:drawing}
\end{figure}

In the proposed representation, the prediction is an element from a product space
over probability simplices in the form of 
$(p, q^0, q^1, \dotsc, q^M)$, where $p \in \Delta^{(M+1)}$ and
$q^i \in \Delta^{g_i+1}$ for all $i=0,1,\dotsc,M$. Therefore number of activations at the group-wise soft-max layer 
$(M+1) + \sum_{i=0}^{M} (g_i + 1) = 2(M+1) + \sum_{i=0}^Mg_i = 2(M+1)+N $. Thus we require only extra $2(M+1)$ activations compared to the conventional semantic segmentation with cross-entropy loss. Further, we use only a small number of groups (for example, our dataset consists of five groups and consequently requires $12$ more activations), thus 
proposed method comprises approximately similar number of parameters [see Fig.~\ref{fig:drawing}].

\section{Evaluation}
\label{sec:evaluation}

%\paragraph{\bf Parameter Settings}

The proposed segmentation method is evaluated with a standard semantic segmentation architecture (described in sec.~\ref{sec:architecture}) on a newly developed dataset, augmented from SUNCG~\cite{song2017semantic} (details are in the following section). 
The loss $\cL$ (eqn. \eqref{eq:loss}) is minimized using ADAM with a mini-batch of size  $25$. 
The weight decay is set to $10^{-5}$. The network is trained for $100$ epochs with an initial learning rate $0.001$ which 
is gradually decreased by a factor of $10$ after every $10$ epochs. A copy of same set of parameters is utilized for the baseline method. 
The architecture is implemented in \texttt{Python} and trained with \texttt{Tensorflow}\footnote{\href{https://www.tensorflow.org/}{https://www.tensorflow.org/}} on a desktop equipped with a NVIDIA Titan X GPU and an Intel CPU of $3.10GHz$.

\begin{table}
\caption{The partition of objects into different Groups of {\bf SUNCG}~\cite{song2017semantic} }
\scriptsize \centering   
\begin{tabular}{ p{2.9cm} |@{\hspace{0.2cm}} l}  
 \toprule 
 Groups $\cG_i$ & Objects $\cO_j$ in the group\\
 \midrule 
 Group $\cG_0$: Background & 'ceiling', 'floor',	'wall',	'window', 'door'  \\ 
 Group $\cG_1$: Chair-like & 'chair', 'table$\_$and$\_$chair',	'trash$\_$can', 'toilet'	\\ 
 Group $\cG_2$: Table-like & 'table',	'side$\_$table',	'bookshelf',  'desk' \\ 
 Group $\cG_3$: Big Objects & 'bed', 'kitchen$\_$cabinet',	'bathtub', 'mirror',   \\ 
 & 	'closets$\_$cabinets',  'dont$\_$care', 'sofa' \\ 
 Group $\cG_4$: Small Objects & 'lamp',	'computer', 'music',	'gym',	'pillow',  	\\ 
 & 'household$\_$appliance', 'kitchen$\_$appliance', \\ 
 & 'pets',	'car',	'plants', 'pool',	'recreation',  \\ 
 & 'night$\_$stand', 'shower', 'tvs',	 'sink' \\ 
 \midrule 
 \end{tabular} 
 \begin{tabular}{c}
In addition, as discussed before, a ``void'' category is inserted 
in each group \\ indicating absence of the object categories in the group while rendering. \\ 
 \bottomrule  
  \end{tabular} 
\label{tab:groupSUNCG}
\end{table}

\subsection{Datasets }\label{suncgdatasets}
We could not find any suitable dataset for the current task. As described above, the amodal semantic segmentation dataset~\cite{zhu2017semantic}
does not serve our purpose. Note that we require group-wise semantic labels
for the evaluation, thus the following datasets are leveraged in this work: %is explored and tailored to the current task. 
\subsubsection{{\bf SUNCG}~\cite{song2017semantic}}
\begin{figure*}
\centering \small 
\begin{tabular}{@{\hspace{-0.05em}}l@{\hspace{1em}}c@{\hspace{0.05em}}c@{\hspace{0.05em}}c@{\hspace{0.05em}}c@{\hspace{0.05em}}c@{\hspace{0.05em}}c@{\hspace{0.05em}}c} \vspace*{1pt} 
& (a) {Depth} & (b) {Sem-Seg} & (c) $\cG_0$  & (d) $\cG_1$ & (e) $\cG_2$ & (f) $\cG_3$  & (g) $\cG_4$  \\ \vspace*{-2pt}
%\begin{picture}(1,25)
%  \put(0,-5){\rotatebox{90}{~~~~~\color{blue}{Samp. 1}}}
%\end{picture} & 
%{\includegraphics[width=0.125\textwidth]{sample/A_924_0007.png}} & 
%{\includegraphics[width=0.125\textwidth]{sample/B_924_0007_label.png}} & 
%{\includegraphics[width=0.125\textwidth]{sample/B_924_0007_label1.png}} & 
%{\includegraphics[width=0.125\textwidth]{sample/B_924_0007_label2.png}} & 
%{\includegraphics[width=0.125\textwidth]{sample/B_924_0007_label3.png}} & 
%{\includegraphics[width=0.125\textwidth]{sample/B_924_0007_label4.png}} & 
%{\includegraphics[width=0.125\textwidth]{sample/B_924_0007_label5.png}} \\ 
%\vspace*{-2pt}
\begin{picture}(1,25)
  \put(0,-5){\rotatebox{90}{~~~~~\color{blue}{Sample 1}}}
\end{picture} & 
{\includegraphics[width=0.125\textwidth]{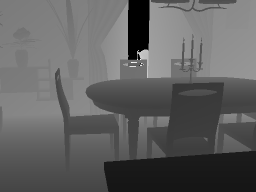}} & 
{\includegraphics[width=0.125\textwidth]{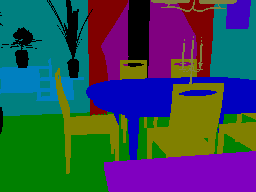}} & 
{\includegraphics[width=0.125\textwidth]{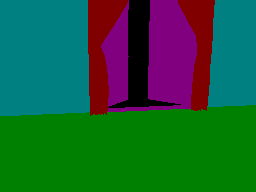}} & 
{\includegraphics[width=0.125\textwidth]{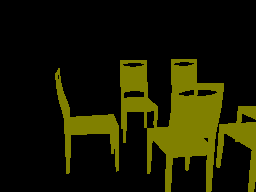}} & 
{\includegraphics[width=0.125\textwidth]{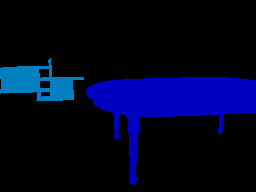}} & 
{\includegraphics[width=0.125\textwidth]{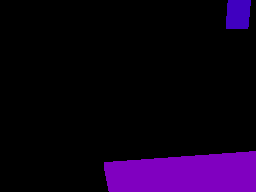}} & 
{\includegraphics[width=0.125\textwidth]{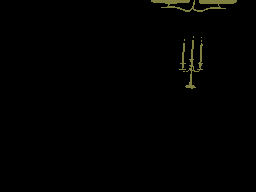}} \\ 
\begin{picture}(1,25)
  \put(0, 0){\rotatebox{90}{~~~~~\color{blue}{Sample 2}}}
\end{picture} & 
{\includegraphics[width=0.125\textwidth]{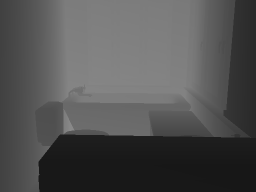}} & 
{\includegraphics[width=0.125\textwidth]{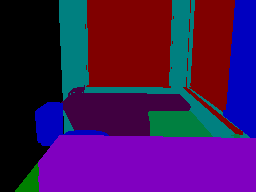}} & 
{\includegraphics[width=0.125\textwidth]{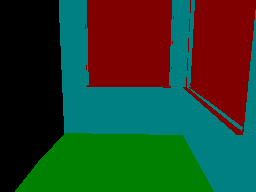}} & 
{\includegraphics[width=0.125\textwidth]{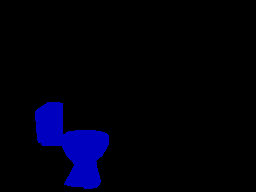}} & 
{\includegraphics[width=0.125\textwidth]{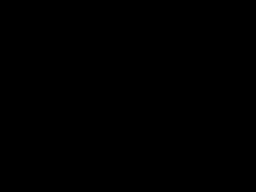}} & 
{\includegraphics[width=0.125\textwidth]{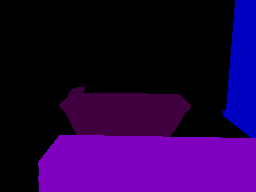}} & 
{\includegraphics[width=0.125\textwidth]{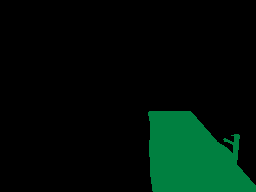}} %\\  \vspace*{-2pt}
\end{tabular} 
 \caption{Examples of images of our synthesized dataset tailored from SUNCG Datasets~\cite{song2017semantic}. Note that given a single depth image (a), the task is to estimate the semantic labels of the different objects present in the scene, along with predict individual groups $[\cG_0,\;\cG_1,\;\cG_2,\;\ldots,\cG_M]$ of different objects.  See sec.~\ref{suncgdatasets} for the details. The ``void'' category $\varnothing$ is marked by black pixels.} % The details of the color coding of the different object categories can be found in the supplementary material.} 
  \label{fig:SUNCGSample}
\end{figure*}
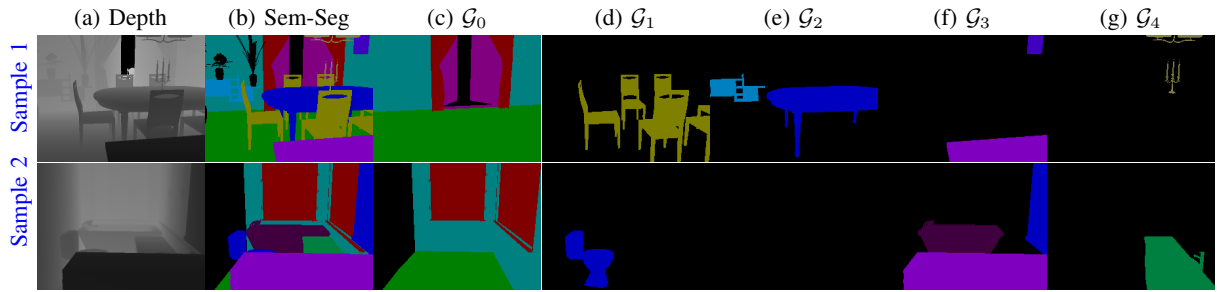

\begin{itemize}[noitemsep,nolistsep,leftmargin=5.5mm,topsep=1ex,partopsep=1ex]  
\item  It is a large-scale dataset containing $6,220$ synthetic 3D models of indoor scenes. % The dataset also provides a large number of RGB and depth image pairs which are not utilized in this work. 
Further, it provides a toolbox \footnote{\href{https://github.com/shurans/SUNCGtoolbox}{https://github.com/shurans/SUNCGtoolbox}} to select camera poses at informative locations of the 3D scene. The toolbox generates $22,742$ distinct camera poses which are further  
augmented $20$ times with random traversal at the nearby poses. The random traversals are considered as Gaussians with standard deviations $1$m  and $15^\circ$ for translations and rotation respectively. 
The camera poses are further refined to get $159,837$ distinct camera poses. We incorporate following heuristics for the refinement, \emph{i.e.},  remove non-informative camera poses if     
\begin{itemize}
\item the scene does not contain any other objects excluding background object categories inside the viewing frustum of the camera pose,  
\item there is substantial portion ($40\%$) of an object placed very close to the camera (within $1$m of the camera center) and inside the viewing frustum, 
\item more than $40\%$ of the ground-truth pixel labels are assigned to `dont$\_$care' object category,  etc.    
\end{itemize}

\item For a given scene (a synthetic 3D model) and a camera pose, the toolbox also provides a rendering engine that generates depth images and semantic labels of $36$ object categories. The depth image and semantic label image pair can be used as a ground-truth training image pair to train a conventional direct semantic segmentation method.  
\item To generate ground-truths for the proposed group-wise semantic segmentation, the object categories are further grouped intuitively into $5$ different groups as described in Table \ref{tab:groupSUNCG} and are rendered individually. Note that no optimal strategy for grouping is utilized in this work. The ground-truth occlusions among the different groups are generated as follows---remove all the objects in the scene 
and place only the objects present in each group one at a time. The rendering engine is then applied to each group 
to generate the semantic labels of the objects present in the group. Examples of the ground-truth dataset can be found in Fig.~\ref{fig:SUNCGSample}. 

\item A similar strategy is applied to $155$ synthetic test 3D models to generate $873$ ground-truth test images. Note that we used the same partition of the training and test 3D models provided by the dataset. 
\end{itemize}

\subsubsection{{\bf Cityscape}~\cite{cordts2016cityscapes}}
\begin{itemize}[noitemsep,nolistsep,leftmargin=5.5mm,topsep=-1ex,partopsep=-1ex]   
\item A real large-scale dataset for semantic segmentation on real driving scenarios. The original dataset does not include amodal perception
 (i.e., separate visible / occluded labels for each pixel). 

\item To evaluate the proposed method in this dataset, the objects are divided into $3$ different groups as described in Table~\ref{tab:groupCityscapes}. Similar to SUNCG dataset, we follow an intuitive strategy to group the objects.    
However in this scenario, it is hard to generate a real dataset of semantic labels of the occluded objects. 
A similar strategy of \cite{li2016amodal} is adapted to generate the training dataset. We randomly duplicate mobile objects 
on relevant locations---for example, `car',  `truck',  `bus' on the `road' and `person', `rider' on the `sidewalk'---and corresponding visible (newly placed object label) and occluded (original semantic label) locations are utilized for training. During training the network hallucinates the 
occluded labels which is matched against the ground-truth occluded labels (ignored if unavailable) and penalizes any miss-predictions. 
Note that during testing our method only takes a real rgb image and generate semantic visible layers of each 
group (described in Table~\ref{tab:groupCityscapes}). % which intern can estimate  full {Sem-Seg}.  

\item  The detailed results are furnished in sec.~\ref{sec:resultscityscapes}. 
\end{itemize}

\begin{table}
\caption{The partition of object categories of {\bf Cityscape} dataset~\cite{cordts2016cityscapes}}
\centering
\scriptsize 
\begin{tabular}{ p{3.1cm} |@{\hspace{0.2cm}} l}  
 \toprule 
 Groups & Objects in the group\\
 \midrule 
 Group $\cG_0$: Background & 'road', 'sidewalk',	'wall',	'building', 'sky',  \\
 & 'terrain', 'fence',  'vegetation' \\ 
 Group $\cG_1$: Traffic Objects & 'pole', 'traffic light',	'traffic sign'	\\ 
 Group $\cG_2$: Mobile Objects & 'person', 'rider', 'car',  'truck',  'bus',  \\
 & 'motocycle', 'train', 'bicycle'\\ 
 \bottomrule  
\end{tabular}
\label{tab:groupCityscapes}
\end{table}

\subsection{Baseline Methods} The evaluation is conducted amongst following baselines: 
\begin{itemize}[noitemsep,nolistsep,leftmargin=5.5mm,topsep=1ex,partopsep=1ex]  
\item The conventional semantic segmentation network (U-net~\cite{ronneberger2015u} architecture) with cross-entropy loss named as \emph{\bf Direct {Semantic Segmentation} (DSS)}. The final semantic segmentation result is abbreviated as {Sem-Seg}.  
\item A similar architecture is utilized for the proposed group-wise segmentation--instead of direct pixel-wise classification  a group-wise classification is introduced as described in sec. \ref{sec:grouping}. We \emph{abbreviate the proposed method} as 
\emph{\bf Grouped {Semantic Segmentation} (GSS)}. 
\end{itemize}
Note that in this work, we emphasize on  group-wise (\emph{e.g.}, ``Chair-like'') segmentation for the task of semantic labeling of the occluded objects compared to conventional semantic segmentation methods. Therefore the proposed method is not evaluated against different types of semantic segmentation architectures. Further, our group-wise model can be 
adapted to any of the existing sophisticated architectures.

\begin{figure*}
\centering \small 
\begin{tabular}{@{\hspace{-0.05em}}l@{\hspace{1em}}c@{\hspace{0.05em}}c@{\hspace{0.05em}}c@{\hspace{0.05em}}c@{\hspace{0.05em}}c@{\hspace{0.05em}}c@{\hspace{0.05em}}c} 
& (a) {Input}  & (b) $\cG_0$  & (c) $\cG_1$ & (d) $\cG_2$ & (e) $\cG_3$  & (f) $\cG_4$  & (g) {Sem-Seg} \vspace*{2pt} \\ \toprule  \vspace*{-2pt} 
\begin{picture}(1,25)
  \put(0, -5){\rotatebox{90}{~~~~~\color{red}{ DSS}}}
\end{picture} & 
{\includegraphics[width=0.125\textwidth]{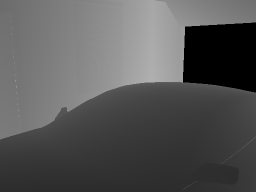}}  & 
{\includegraphics[width=0.125\textwidth]{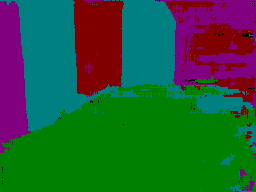}} & 
{\includegraphics[width=0.125\textwidth]{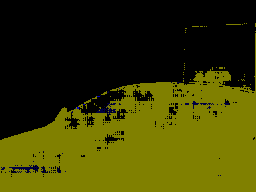}} & 
{\includegraphics[width=0.125\textwidth]{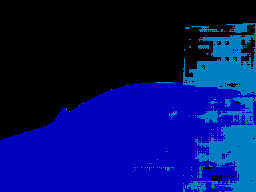}} & 
{\includegraphics[width=0.125\textwidth]{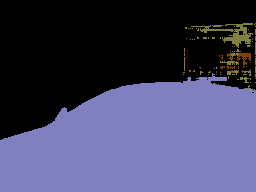}} & 
{\includegraphics[width=0.125\textwidth]{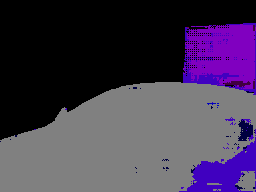}} & 
{\includegraphics[width=0.125\textwidth]{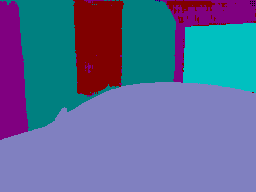}} \vspace*{0pt}\\ 

\begin{picture}(1,25)
  \put(0, -10){\rotatebox{90}{~~~~~\color{blue}{ GSS}}}
\end{picture} & 
  & 
{\includegraphics[width=0.125\textwidth]{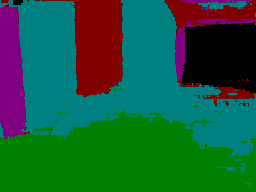}} & 
{\includegraphics[width=0.125\textwidth]{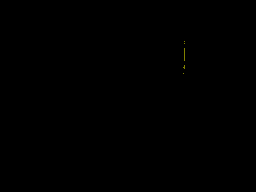}} & 
{\includegraphics[width=0.125\textwidth]{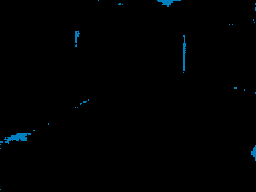}} & 
{\includegraphics[width=0.125\textwidth]{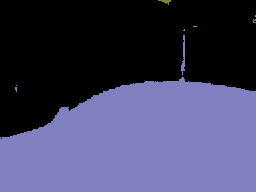}} & 
{\includegraphics[width=0.125\textwidth]{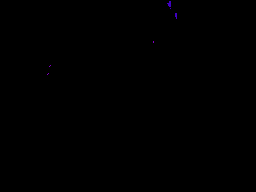}} & 
{\includegraphics[width=0.125\textwidth]{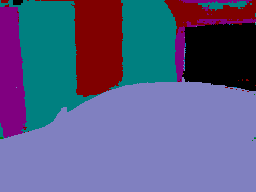}} \vspace*{-2pt}\\ 

\begin{picture}(1,25)
  \put(0, -5){\rotatebox{90}{~~~~~\color{purple}{ GT}}}
\end{picture} & 
  & 
{\includegraphics[width=0.125\textwidth]{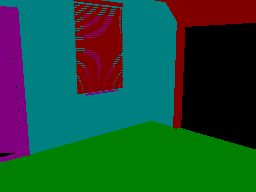}} & 
{\includegraphics[width=0.125\textwidth]{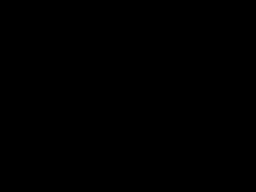}} & 
{\includegraphics[width=0.125\textwidth]{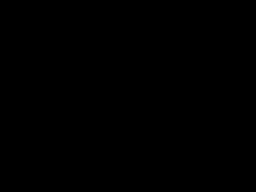}} & 
{\includegraphics[width=0.125\textwidth]{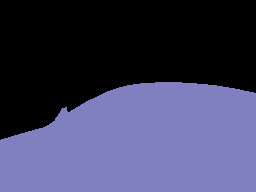}} & 
{\includegraphics[width=0.125\textwidth]{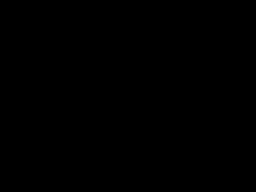}} & 
{\includegraphics[width=0.125\textwidth]{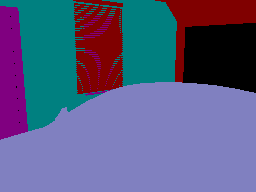}} \vspace*{-2pt} \\ \midrule  \vspace*{-10pt} \\ \vspace*{-2pt}

\begin{picture}(1,25)
  \put(0, 0){\rotatebox{90}{~~~~~\color{red}{ DSS}}}
\end{picture} & 
{\includegraphics[width=0.125\textwidth]{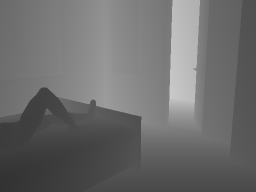}} & 
{\includegraphics[width=0.125\textwidth]{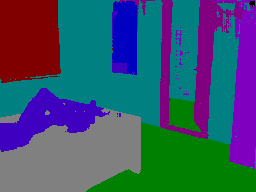}} & 
{\includegraphics[width=0.125\textwidth]{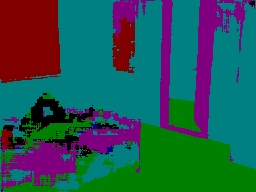}} & 
{\includegraphics[width=0.125\textwidth]{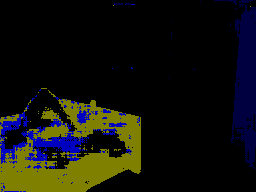}} & 
{\includegraphics[width=0.125\textwidth]{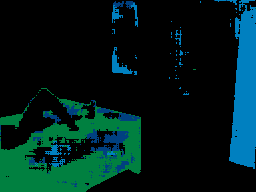}} & 
{\includegraphics[width=0.125\textwidth]{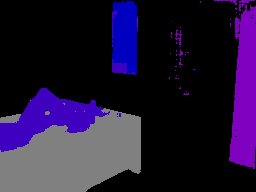}} & 
{\includegraphics[width=0.125\textwidth]{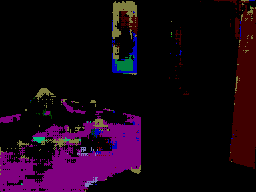}} \vspace*{0pt} \\ 

\begin{picture}(1,25)
  \put(0, 0){\rotatebox{90}{~~~~~\color{blue}{ GSS}}}
\end{picture} & 
 & 
{\includegraphics[width=0.125\textwidth]{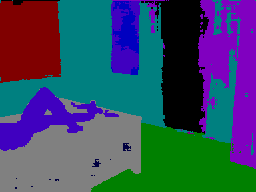}} & 
{\includegraphics[width=0.125\textwidth]{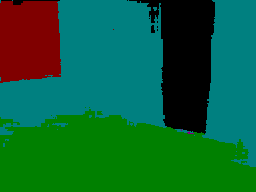}} & 
{\includegraphics[width=0.125\textwidth]{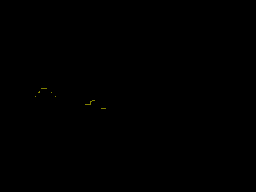}} & 
{\includegraphics[width=0.125\textwidth]{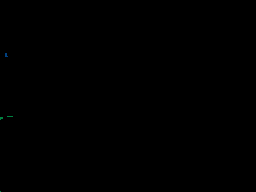}} & 
{\includegraphics[width=0.125\textwidth]{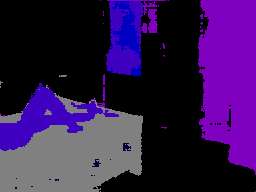}} & 
{\includegraphics[width=0.125\textwidth]{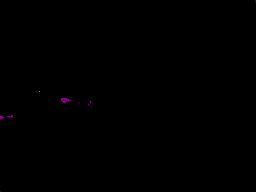}} \vspace*{-2pt} \\

\begin{picture}(1,25)
  \put(0, 0){\rotatebox{90}{~~~~~\color{blue}{ GT}}}
\end{picture} & 
 & 
{\includegraphics[width=0.125\textwidth]{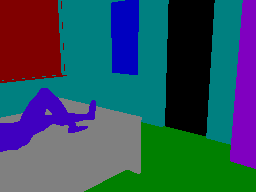}} & 
{\includegraphics[width=0.125\textwidth]{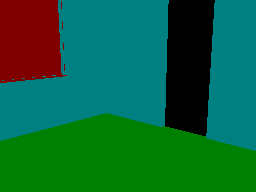}} & 
{\includegraphics[width=0.125\textwidth]{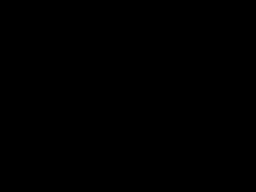}} & 
{\includegraphics[width=0.125\textwidth]{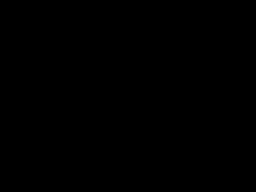}} & 
{\includegraphics[width=0.125\textwidth]{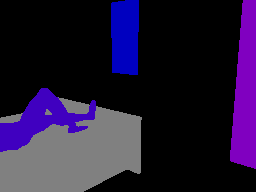}} & 
{\includegraphics[width=0.125\textwidth]{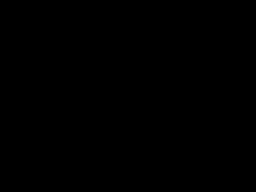}} \vspace*{-2pt} \\ \midrule  \vspace*{-10pt} \\ \vspace*{-2pt}

\begin{picture}(1,25)
  \put(0, -5){\rotatebox{90}{~~~~~\color{red}{ DSS}}}
\end{picture} & 
{\includegraphics[width=0.125\textwidth]{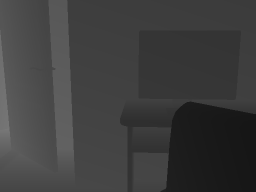}}  & 
{\includegraphics[width=0.125\textwidth]{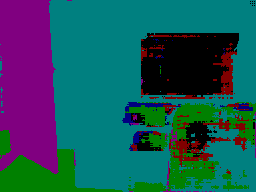}} & 
{\includegraphics[width=0.125\textwidth]{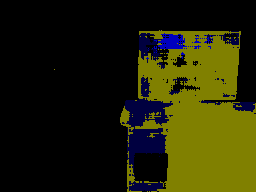}} & 
{\includegraphics[width=0.125\textwidth]{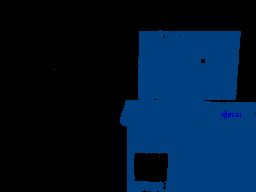}} & 
{\includegraphics[width=0.125\textwidth]{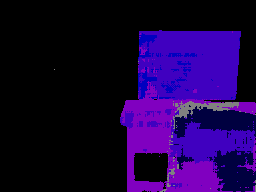}} & 
{\includegraphics[width=0.125\textwidth]{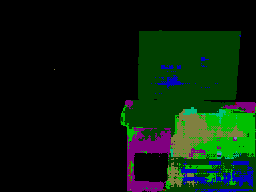}} & 
{\includegraphics[width=0.125\textwidth]{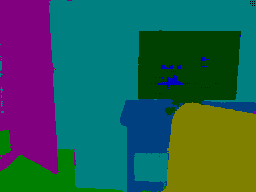}} \vspace*{0pt} \\ 

\begin{picture}(1,25)
  \put(0, -10){\rotatebox{90}{~~~~~\color{blue}{ GSS}}}
\end{picture} & 
  & 
{\includegraphics[width=0.125\textwidth]{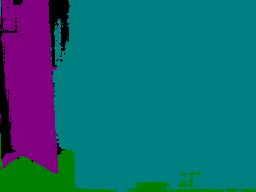}} & 
{\includegraphics[width=0.125\textwidth]{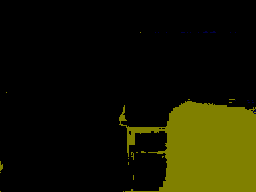}} & 
{\includegraphics[width=0.125\textwidth]{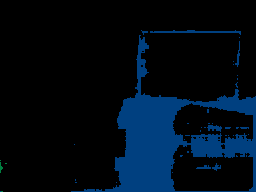}} & 
{\includegraphics[width=0.125\textwidth]{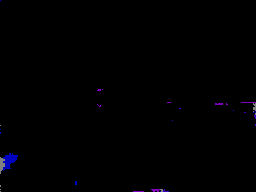}} & 
{\includegraphics[width=0.125\textwidth]{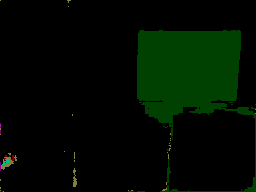}} & 
{\includegraphics[width=0.125\textwidth]{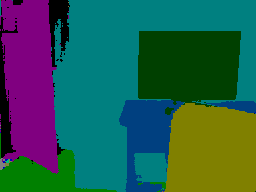}} \vspace*{-2pt} \\
\begin{picture}(1,25)
  \put(0, -5){\rotatebox{90}{~~~~~\color{purple}{ GT}}}
\end{picture} & 
& 
{\includegraphics[width=0.125\textwidth]{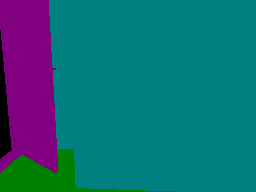}} & 
{\includegraphics[width=0.125\textwidth]{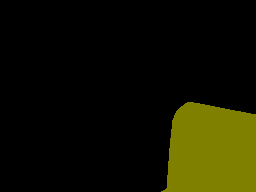}} & 
{\includegraphics[width=0.125\textwidth]{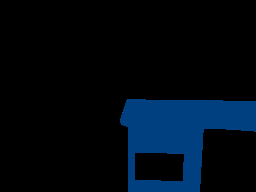}} & 
{\includegraphics[width=0.125\textwidth]{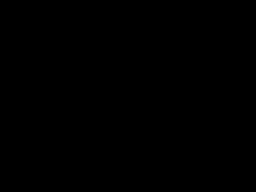}} & 
{\includegraphics[width=0.125\textwidth]{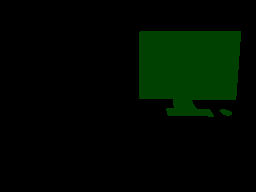}}  & 
{\includegraphics[width=0.125\textwidth]{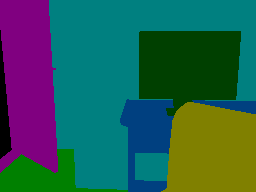}} \vspace*{-2pt} \\ \midrule  \vspace*{-10pt} \\ \vspace*{-2pt} 

\begin{picture}(1,25)
  \put(0, -5){\rotatebox{90}{~~~~~\color{red}{ DSS}}}
\end{picture} & 
{\includegraphics[width=0.125\textwidth]{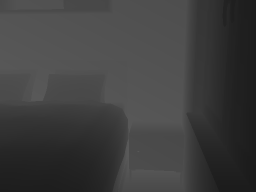}}  & 
{\includegraphics[width=0.125\textwidth]{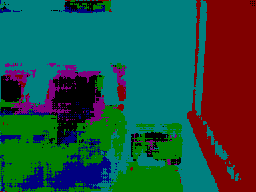}} & 
{\includegraphics[width=0.125\textwidth]{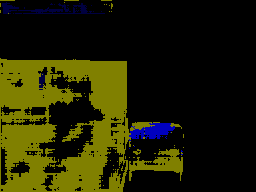}} & 
{\includegraphics[width=0.125\textwidth]{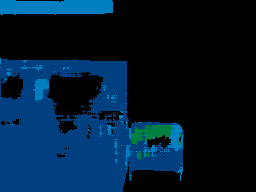}} & 
{\includegraphics[width=0.125\textwidth]{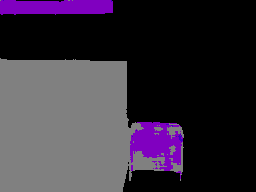}} & 
{\includegraphics[width=0.125\textwidth]{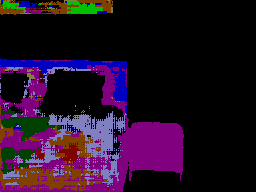}} & 
{\includegraphics[width=0.125\textwidth]{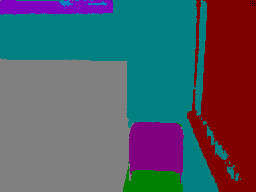}} \vspace*{0pt} \\ 

\begin{picture}(1,25)
  \put(0, -10){\rotatebox{90}{~~~~~\color{blue}{ GSS}}}
\end{picture} & 
& 
{\includegraphics[width=0.125\textwidth]{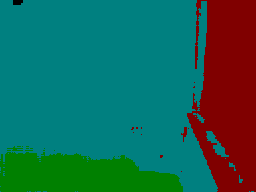}} & 
{\includegraphics[width=0.125\textwidth]{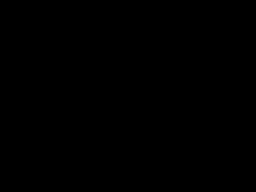}} & 
{\includegraphics[width=0.125\textwidth]{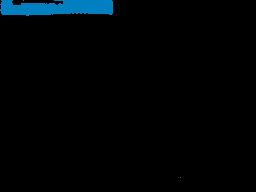}} & 
{\includegraphics[width=0.125\textwidth]{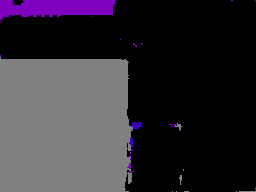}} & 
{\includegraphics[width=0.125\textwidth]{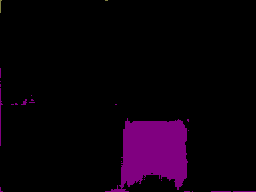}}  & 
{\includegraphics[width=0.125\textwidth]{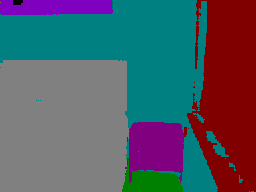}} \vspace*{-2pt} \\ 
\begin{picture}(1,25)
  \put(0, -5){\rotatebox{90}{~~~~~\color{purple}{ GT}}}
\end{picture} & 
& 
{\includegraphics[width=0.125\textwidth]{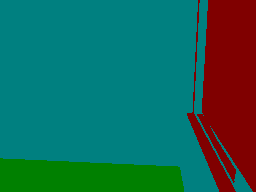}} & 
{\includegraphics[width=0.125\textwidth]{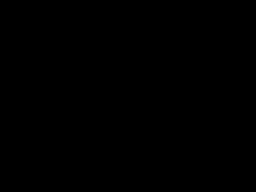}} & 
{\includegraphics[width=0.125\textwidth]{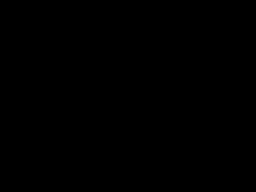}} & 
{\includegraphics[width=0.125\textwidth]{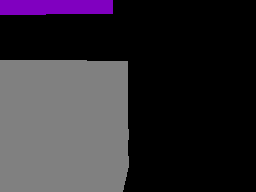}} & 
{\includegraphics[width=0.125\textwidth]{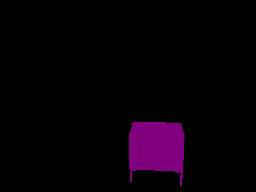}}  & 
{\includegraphics[width=0.125\textwidth]{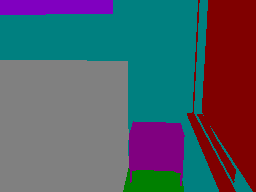}}  \vspace*{-2pt} \\ \bottomrule  \vspace*{-10pt} 

\end{tabular} 

 \caption{Qualitative results on SUNCG Datasets~\cite{song2017semantic}. Note that the floor parts are hallucinated almost perfectly. See sec.~\ref{sec:qualitative} for more details. } 
  \label{fig:SUNCGResults}
\end{figure*}

\subsection{Quantitative Evaluation}\label{sec:quantitative}
Many evaluation criteria have been proposed in the literature to assess the quality of the semantic segmentation methods. 
In this work, we adapt existing matrices and tailor in the following way to tackle occlusions 
\begin{itemize}[noitemsep,nolistsep,leftmargin=5.5mm,topsep=1ex,partopsep=1ex]
\item {\bf Visible Pixel Accuracy ($\text{PA}^{vis}$):} It is the ratio between properly classified (visible) pixels and the total number of them. Let $\Omega_c^{vis}$ and $\widehat{\Omega}_c^{vis}$ be the set of all pixels actually visible and predicted visible with class-label $c$ respectively. 
\begin{equation}
\text{PA}^{vis} = \frac{\sum_{c=1}^N |\Omega_c^{vis} \cap \widehat{\Omega}_c^{vis}| }{|\Omega|}
\end{equation} 

\item {\bf Visible Mean Intersection over Union ($\text{MIoU}^{vis}$):} It is the conventional metric for segmentation purposes and redefined similarly. It is the ratio between the intersection and the union of two sets---the ground truth visible and our predicted visible. It can be written as  
\begin{equation}
\text{MIoU}^{vis} = \frac{1}{N}\sum_{c=1}^N\frac{|\Omega_c^{vis} \cap \widehat{\Omega}_c^{vis}|}{|\Omega_c^{vis} \cup \widehat{\Omega}_c^{vis}|}
\end{equation}   
The IoU is computed on a per class basis and then averaged. 
Note that $\text{PA}^{vis}$ and $\text{MIoU}^{vis}$ are the same as conventional pixel accuracy and mean intersection over union used in the semantic segmentation literature. 

\item {\bf Present Pixel Accuracy ($\text{PA}^{pres}$):} It is defined in a similar way to $\text{PA}^{vis}$ where in this case the numerator is calculated over the pixels where the object is present (visible / occluded). Let $\Omega_c^{pres}$ and $\widehat{\Omega}_c^{pres}$ be the set of all pixels actually present (visible / occluded) and predicted present with class-label $c$ respectively. 

\begin{equation}
\text{PA}^{pres} = \frac{\sum_{c=1}^N|\Omega_c^{pres} \cap \widehat{\Omega}_c^{pres}| }{|\Omega|}
\end{equation} 

\item {\bf Present Mean Intersection over Union ($\text{MIoU}^{pres}$):} It is defined in the similar fashion 
\begin{equation}
\text{MIoU}^{pres} = \frac{1}{N}\sum_{c=1}^N\frac{|\Omega_c^{pres} \cap \widehat{\Omega}_c^{pres}|}{|\Omega_c^{pres} \cup \widehat{\Omega}_c^{pres}|}
\end{equation} 

%
%\item {\bf Mean Pixel Accuracy (MPA):} the ratio of correct pixels computed in a per-class basis and then averaged over the total number of classes. 
%\begin{equation}
%MPA = \frac{1}{k+1}\sum_{i=0}^{k}\frac{p_{ii}}{\sum_{j=0}^{k}p_{ij}}
%\end{equation}
\end{itemize} 
Note that  $(\Omega_c^{vis},\; \Omega_c^{pres})$ are readily available from the ground truth dataset and $(\widehat{\Omega}_c^{vis},\;\widehat{\Omega}_c^{pres})$ are estimated by baseline methods.  
 The conventional DSS estimates $\widehat{\Omega}_c^{vis}$ (set of the pixels predicted with category $c$), however, $\widehat{\Omega}_c^{pres}$ is not readily available. 
It can be obtained by computing indices of the maximum posterior probability of the group (+background) that object category $c$ belongs to i.e., \  
\begin{equation}
\widehat{\Omega}_c^{pres} = \{x | c = \arg\max_{j\in I_i} p_j(x)\}
\end{equation}
where category $c$ belongs to the group $\cG_i$, $p_j(x)$ is the posterior probability of the $j$th class category at pixel $x$, and $I_i$ is the set of indices corresponds to object categories in the set $\cG_0 \cup \cG_i$. The indices of $\cG_0$ are squashed to the ``void'' category $\varnothing$.  If $c$ belongs to the background class $\cG_0$, the index set $I_i$ is chosen from the indices of $\cG_0$ only and no squashing is required in this case.   

\noindent Similarly, the GSS directly estimates $\widehat{\Omega}_c^{pres}$ whereas $\widehat{\Omega}_c^{vis}$ is again not readily available and computed in the following manner: 
\begin{align}
\widehat{\Omega}_c^{vis} &= \{x | c = \cc(\hat{i},\;\hat{j}) \text{~where~}  \hat{i} = \max_i p_i(x) \\ & ~~~~~~~~~~~~~~~~~~~~~~~~\text{~and~} \hat{j} = \max_{j=1,\dotsc,g_i} q^i_j(x) \} \nonumber \\ 
 & = \widehat{\Omega}_{\cG_i}^{vis} \cap \widehat{\Omega}_c^{pres} 
\end{align}
where $\cc(i,j)$ be the reverse mapping yielding the category corresponding to group $i$ and index $j$ and $\widehat{\Omega}_{\cG_i}^{vis}$ are the visible pixels of the group that the category $c$ belongs to. 

The quantitative results are presented in Table~\ref{tab:groupSUNCG2}. The conventional DSS is trained with visible semantic labels [Fig.~\ref{fig:SUNCGSample}(b)] and the proposed GSS method is trained with  visible + occluded [Fig.~\ref{fig:SUNCGSample}(b)-(c)] semantic labels. We observe GSS performs analogously with DSS for the task of conventional semantic segmentation (i.e., visible pixels). However, for the regions where the objects are present (visible / occluded) proposed GSS performs much better. To make it clear that the better performance of GSS is not driven by the inclusion of the ``void'' category, we present results  with / without the ``void'' class separately. Therefore given a fixed network capacity and with the availability of data, the proposed GSS exceeds the conventional method for the task of semantic labeling of the occluded objects. 

\begin{table}\setlength{\tabcolsep}{3pt}
\caption{Quantitative Evaluation on the augmented SUNCG~\cite{song2017semantic}}
\begin{tabular}{ p{1.2cm}|cc|cc|cc}  
 \toprule 
 & \multicolumn{2}{c|}{ Visible } & \multicolumn{2}{|c}{Present Regions} & \multicolumn{2}{|c}{Present Regions} \\ 
 Methods & \multicolumn{2}{c|}{ Regions } & \multicolumn{2}{|c}{with ``void'' $\varnothing$} & \multicolumn{2}{|c}{without ``void'' $\varnothing$} \\ 
 & $\text{PA}^{vis}$ & $\text{MIoU}^{vis}$  & $\text{PA}^{pres}$ & $\text{MIoU}^{pres}$  & $\text{PA}^{pres}$ & $\text{MIoU}^{pres}$ \\
 \midrule 
 DSS & $ \bf{0.850} $ & $\bf{0.575}$  & $ 0.715 $ & $0.268$   & $ 0.755 $ & $0.594$   
 \\ 
 GSS & $ 0.833 $ & $0.551$  & $\bf{ 0.900} $ & $ \bf{0.470} $  & $ \bf{0.861} $ & $ \bf{0.645} $ \\ 
 \bottomrule  
\end{tabular}
\label{tab:groupSUNCG2}
\end{table}

\subsection{Qualitative Evaluation}\label{sec:qualitative}
The qualitative comparison of the selected baseline methods on test dataset is displayed in Fig.~\ref{fig:SUNCGResults}. The  methods 
take the depth image [Fig.~\ref{fig:SUNCGResults}(a)] as an input and predicts the semantic segmentation [Fig.~\ref{fig:SUNCGResults}(b)] as an output. Further, the proposed GSS is also computed as described in the above and displayed in Fig.~\ref{fig:SUNCGResults}(c)-(g). %The conventional DSS, proposed 
%GSS and ground-truths are marked with {\color{red}{DSS}}, {\color{blue}{GSS}} and {\color{purple}{GT}} respectively.  
 We observe that conventional 
{Sem-Seg} could hallucinate the occluded objects in some cases, however, the network finds difficulty estimating ``void'' class which is essential to model the correct occlusion. For example, in the first row of Fig.~\ref{fig:SUNCGResults}, the object class ``Bed'' covers / occlude the entire floor and the conventional DSS could 
able to hallucinate the occluded floor, however, it further hallucinates other objects categories present in the other groups. 

The proposed GSS performs consistently well on task of semantic segmentation of the visible and the occluded objects. Unlike the DSS, we do not observe any difficulty estimating the ``void'' category.  The estimated background also looks much cleaner.   Note that the examples are not cherry-picked and are quite random from the test data. % A detailed results can be found in the supplementary material. 

\subsection{Qualitative Evaluation on {Cityscape}~\cite{cordts2016cityscapes}}
\label{sec:resultscityscapes}
\begin{figure*}
\centering \small 
\begin{tabular}{@{\hspace{-0.05em}}l@{\hspace{1em}}c@{\hspace{0.05em}}c@{\hspace{0.05em}}c} 
& (a) {Input}  & (b) $\cG_0$: Background & (c) $\cG_1$: Traffic Objects \\
\begin{picture}(1,25)
  \put(0,5){\rotatebox{90}{~~~~~\color{blue}{ GSS}}}
\end{picture} & 
\begin{tikzpicture}
    \node[] at (0,0) {\includegraphics[width=0.28\textwidth]{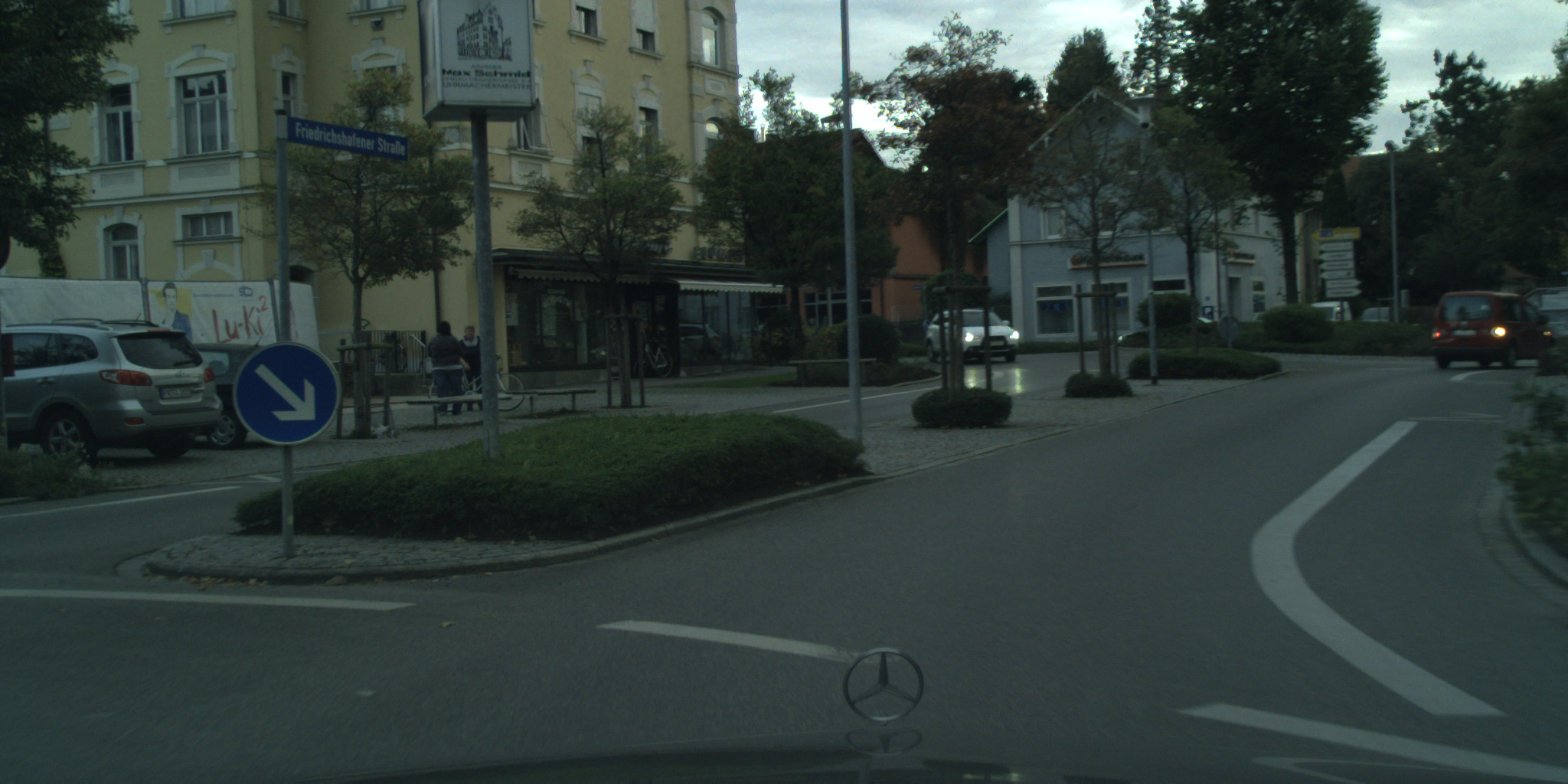}};
    \draw[yellow,ultra thin,rounded corners] (-2.4, -0.7) rectangle (-0.6,0.3);
\end{tikzpicture} & 

\begin{tikzpicture}
    \node[] at (0,0) {\includegraphics[width=0.28\textwidth]{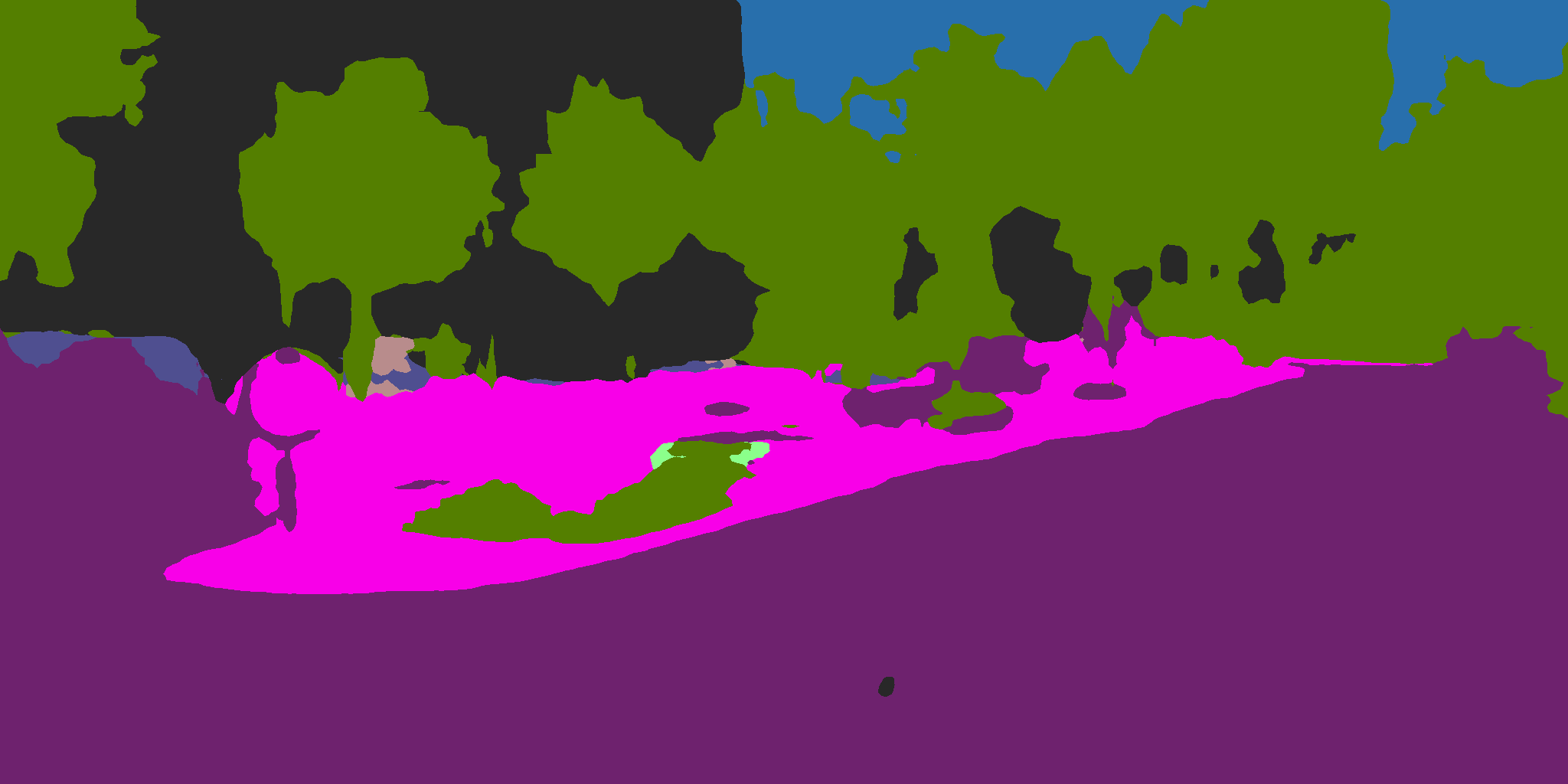}};
    \draw[yellow,ultra thin,rounded corners] (-2.4, -0.7) rectangle (-0.6,0.3);
\end{tikzpicture} & 
\begin{tikzpicture}
    \node[] at (0,0) {\includegraphics[width=0.28\textwidth]{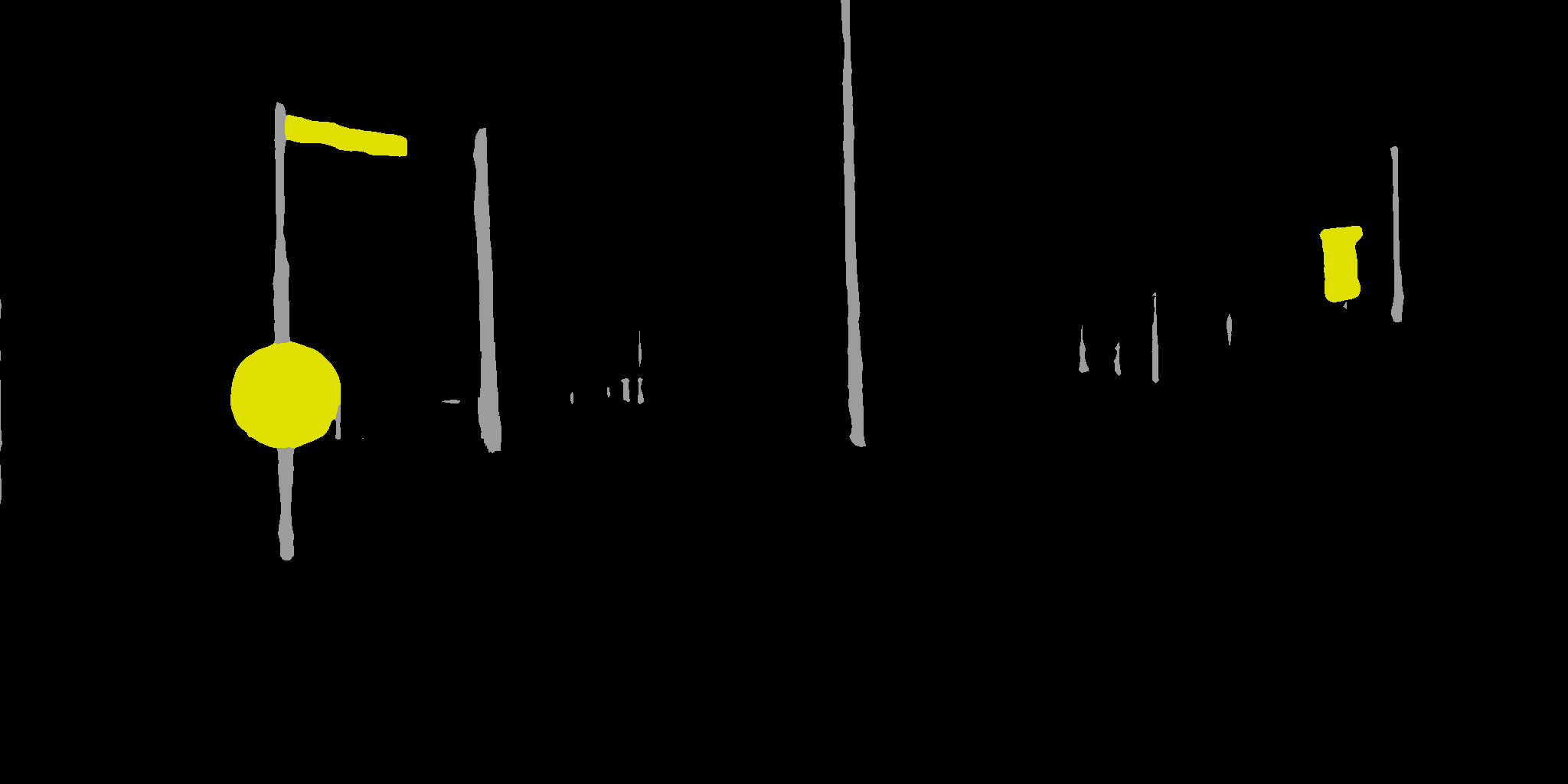}};
    \draw[yellow,ultra thin,rounded corners] (-2.4, -0.7) rectangle (-0.6,0.3);
\end{tikzpicture} \\

 & &  (d) $\cG_2$: Mobile Objects & (e) {Sem-Seg} \\ 

 & & \begin{tikzpicture}
    \node[] at (0,0) {\includegraphics[width=0.28\textwidth]{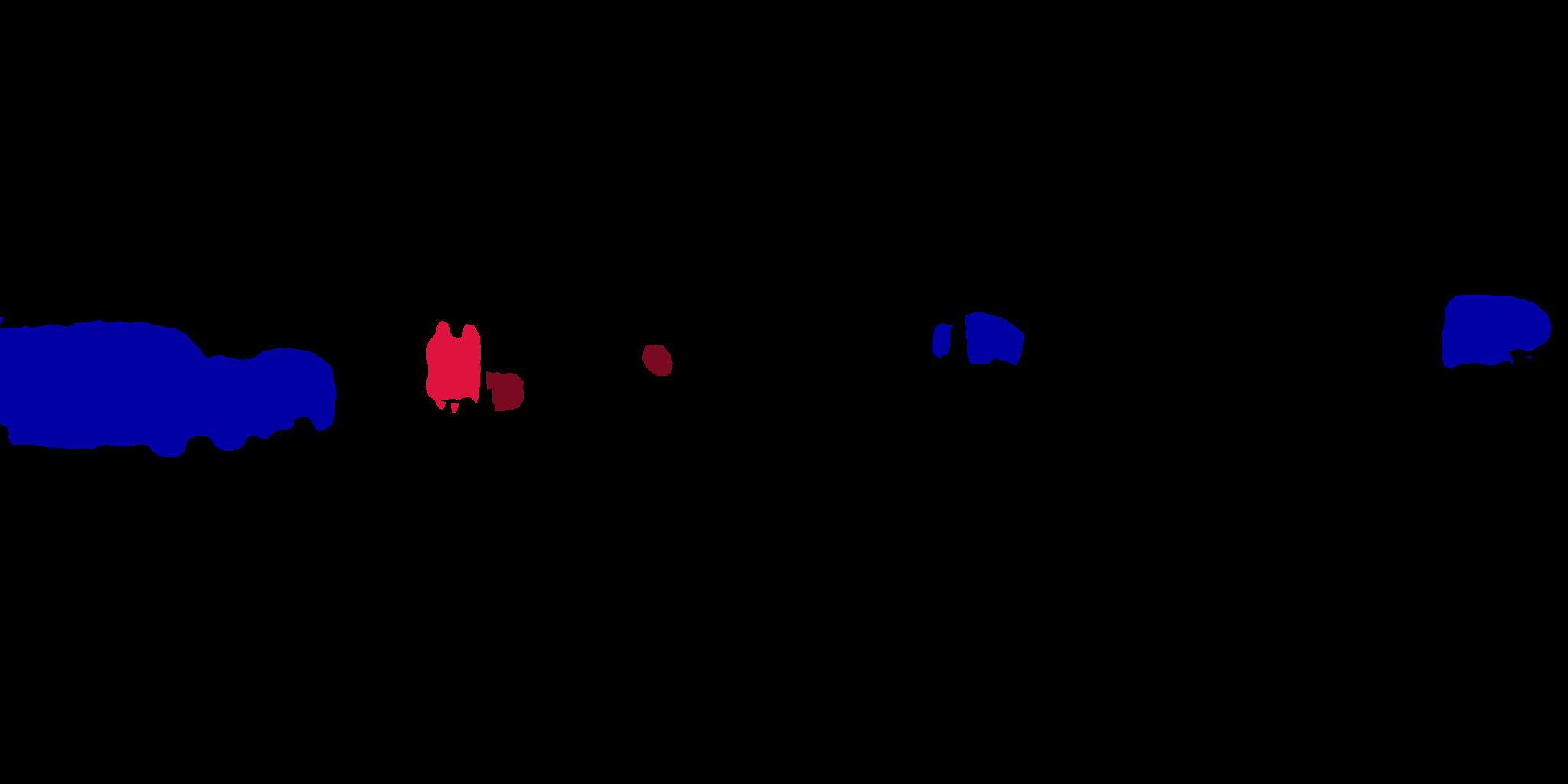}};
    \draw[yellow,ultra thin,rounded corners] (-2.4, -0.7) rectangle (-0.6,0.3);
\end{tikzpicture} & 

\begin{tikzpicture}
    \node[] at (0,0) {\includegraphics[width=0.28\textwidth]{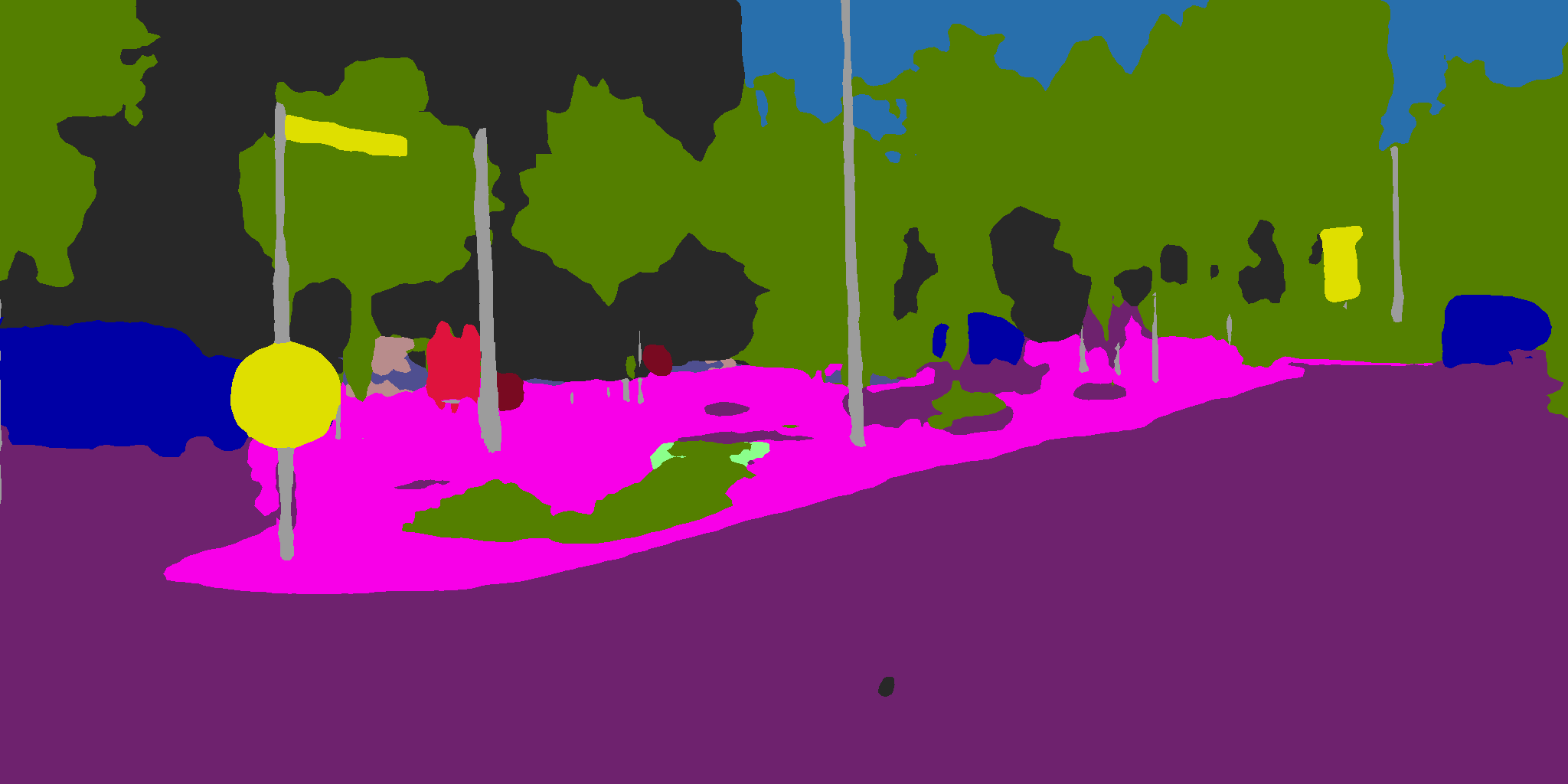}};
    \draw[yellow,ultra thin,rounded corners] (-2.4, -0.7) rectangle (-0.6,0.3);
\end{tikzpicture} \\ 

% {\includegraphics[width=0.31\textwidth]{PSPNet/test2.png}} & 
%{\includegraphics[width=0.31\textwidth]{PSPNet/test2results4.png}} & 
%{\includegraphics[width=0.31\textwidth]{PSPNet/test2results1.png}} \\
% & & (d) $\cG_1$ & (e) $\cG_2$  \\ 
% & & {\includegraphics[width=0.31\textwidth]{PSPNet/test2results2.png}} & 
% {\includegraphics[width=0.31\textwidth]{PSPNet/test2results3.png}} \\

& Cropped--(a) {Input}  & Cropped--(b) $\cG_0$: Background & Cropped--(c) $\cG_1$: Traffic Objects \\
\begin{picture}(1,25)
  \put(0,5){\rotatebox{90}{~~~~~\color{blue}{ GSS}}}
\end{picture} & 
{\includegraphics[width=0.28\textwidth, trim={117 253 1240 418},clip]{PSPNet/test2.png}} & 
{\includegraphics[width=0.28\textwidth, trim={117 253 1240 418},clip]{PSPNet/newtest2results1.png}} & 
{\includegraphics[width=0.28\textwidth, trim={117 253 1240 418},clip]{PSPNet/newtest2results2.png}} \\
 &  & Cropped--(d) $\cG_2$: Mobile Objects &  Cropped--(e)~{Sem-Seg} \\ 
 & & {\includegraphics[width=0.28\textwidth, trim={117 253 1240 418},clip]{PSPNet/newtest2results32.png}} & 
{\includegraphics[width=0.28\textwidth, trim={117 253 1240 418},clip]{PSPNet/newtest2results4.png}} 
\end{tabular}
 \caption{Qualitative results of an image from Cityscape datasets and its cropped version of a region of interest. Notice that the car, hidden behind the traffic signal (marked by the yellow box in (a) and then later zoomed in Cropped-(a)), is fully recovered in the `$\cG_2$: mobile objects' layer (d). However, the traffic signal wrongly cut the pavements into parts (Cropped-(d)) in the `$\cG_0$: Background' layer. } 
 \label{fig:cityscapes}
\end{figure*}

The training datasets for cityscape are augmented by placing the mobile objects randomly around the neighborhood (see sec.~\ref{suncgdatasets} for details). During testing, the network only observes a real rgb image (in contrast to SUNCG~\cite{song2017semantic} that observes a single depth image) and produces the semantic labels for each group. An example result and its cropped version is shown 
in Fig.~\ref{fig:cityscapes}.

\section{Conclusion and future work}
\label{sec:conclusion}
In this work we aim to push the boundaries of assigning semantic categories only to visible objects. 
A group-wise semantic segmentation strategy is proposed, which is capable of predicting semantic labels of the visible objects along with semantic labels of the occluded objects and object parts without any necessity to enrich the network capacity.   
%To our knowledge this work pioneers a simple and efficient framework for predicting semantic labels of occluded object parts in image space.
We develop a synthetic dataset to evaluate our GSS. A standard network architecture trained with proposed grouped semantic loss performs much better than conventional cross-entropy loss for the task of predicting semantic labels of the occluded pixels. The dataset along with the scripts of the current work  will be released to facilitate research towards estimating semantic labels of the occluded objects. 

Currently the training set, leveraged in this work, is purely synthetic, but it
allows strongly supervised training. This limits the applicability of the
proposed method to environments having a suitable synthetic models
available. Hence, future work will address how to incorporate weaker supervision
in this problem formulation in case of a real dataset and also incorporate uncertainty estimation of the occluded regions.

\end{document}